\documentclass[letterpaper]{article} 
\usepackage{aaai23}  
\usepackage{times}  
\usepackage{helvet}  
\usepackage{courier}  
\usepackage[hyphens]{url}  
\usepackage{graphicx} 
\usepackage{natbib}  
\usepackage{caption} 
\usepackage{algorithm}
\usepackage{algorithmic}

\usepackage{newfloat}
\usepackage{listings}
\DeclareCaptionStyle{ruled}{labelfont=normalfont,labelsep=colon,strut=off} 
\floatstyle{ruled}
\newfloat{listing}{tb}{lst}{}
\floatname{listing}{Listing}
\usepackage{amsfonts}
\usepackage{amssymb,amsthm,amsmath}
\usepackage{esint}
\usepackage{isomath}
\usepackage{latexsym}
\usepackage{epic}
\usepackage{epsfig}
\usepackage{multirow}
\usepackage{hhline}
\usepackage{wrapfig}
\usepackage{verbatim}
\usepackage{enumitem}
\usepackage{color}
\usepackage{thmtools}
\usepackage{thm-restate}
\usepackage[labelformat=simple]{subcaption}
\usepackage{mathtools}
\usepackage{graphicx}
\usepackage{tabularx}
\usepackage{tabu}
\usepackage{colortbl}
\usepackage[dvipsnames]{xcolor}         
\usepackage{tikz}
\usetikzlibrary{arrows,automata,positioning}

\newcommand{\R}{\mathbb{R}}

\newcommand{\feature}{x}
\newcommand{\Feature}{\mathcal{X}}
\newcommand{\whittle}{m}
\newcommand{\model}{f} 
\newcommand{\parameter}{P}

\newcommand{\weight}{w}
\newcommand{\tmpstate}{u}
\newcommand{\state}{s}
\newcommand{\nstates}{M}
\newcommand{\State}{\mathcal{S}}
\newcommand{\action}{a}
\newcommand{\Action}{\mathcal{A}}
\newcommand{\reward}{r}
\newcommand{\belief}{b}
\newcommand{\Belief}{\mathcal{B}}
\newcommand{\policy}{\pi}
\newcommand{\trajectories}{\mathcal{T}}
\newcommand{\trajectory}{\tau}
\newcommand{\dataset}{\mathcal{D}}

\theoremstyle{plain}
\newtheorem{theorem}{Theorem}[section]

\theoremstyle{definition}
\newtheorem{definition}[theorem]{Definition}

\theoremstyle{remark}

\title{Scalable Decision-Focused Learning in Restless Multi-Armed Bandits with Application to Maternal and Child Health}
\author{
  Kai Wang\equalcontrib\thanks{Work done during an internship at Google Research.}\textsuperscript{\rm 1},
  Shresth Verma\equalcontrib\textsuperscript{\rm 2},
  Aditya Mate\footnotemark[2]\textsuperscript{\rm 1},
  Sanket Shah\textsuperscript{\rm 1},
  Aparna Taneja\textsuperscript{\rm 2},\\
  Neha Madhiwalla\textsuperscript{\rm 3},
  Aparna Hegde\textsuperscript{\rm 3},
  Milind Tambe\textsuperscript{\rm 1,2}
}
\affiliations{
    \textsuperscript{\rm 1}Harvard University \\
    \textsuperscript{\rm 2}Google Research \\
    \textsuperscript{\rm 3}ARMMAN \\

    \{kaiwang, aditya\_mate, sanketshah\}@g.harvard.edu, \{aparnataneja, milindtambe\}@google.com,\\
    \{neha, aparnahegde\}@armman.org
}

\usepackage{bibentry}

\begin{document}

\maketitle

\begin{abstract}
This paper studies restless multi-armed bandit (RMAB) problems with unknown arm transition dynamics but with known correlated arm features. The goal is to learn a model to predict transition dynamics given features, where the Whittle index policy solves the RMAB problems using predicted transitions. However, prior works often learn the model by maximizing the predictive accuracy instead of final RMAB solution quality, causing a mismatch between training and evaluation objectives. To address this shortcoming, we propose a novel approach for decision-focused learning in RMAB that directly trains the predictive model to maximize the Whittle index solution quality. We present three key contributions: (i) we establish differentiability of the Whittle index policy to support decision-focused learning; (ii) we significantly improve the scalability of decision-focused learning approaches in sequential problems, specifically RMAB problems; (iii) we apply our algorithm to a previously collected  dataset of maternal and child health to demonstrate its performance. Indeed, our algorithm is the first for decision-focused learning in RMAB that scales to real-world problem sizes.
\end{abstract}

\section{Introduction}
Restless multi-armed bandits (RMABs)~\cite{weber1990index,tekin2012online} are composed of a set of heterogeneous arms and a planner who can pull multiple arms under budget constraint at each time step to collect rewards.
Different from the classic stochastic multi-armed bandits~\cite{gittins2011multi,bubeck2012regret}, the state of each arm in an RMAB can change even when the arm is not pulled, where each arm follows a Markovian process to transition between different states with transition probabilities dependent on arms and the pulling decision.
Rewards are associated with different arm states, where the planner's goal is to plan a sequential pulling policy to maximize the total reward received from all arms.
RMABs are commonly used to model sequential scheduling problems where limited resources must be strategically assigned to different tasks sequentially to maximize 
performance. Examples include machine maintenance~\cite{glazebrook2006some}, cognitive radio sensing problem~\cite{bagheri2015restless}, and healthcare~\cite{mate2022field}.

In this paper, we study offline RMAB problems with unknown transition dynamics but with given arm features.
The goal is to learn a mapping from arm features to transition dynamics, which can be used to infer the
dynamics of unseen RMAB problems to plan accordingly. 
Prior works~\cite{mate2022field,sun2018cell} often learn the transition dynamics from the historical pulling data by \textit{maximizing the predictive accuracy}.
However, RMAB performance is evaluated \textit{by its solution quality} derived from the predicted transition dynamics, which leads to a mismatch in the training objective and the evaluation objective.
Previously, decision-focused learning~\cite{wilder2019melding} has been proposed to directly optimize the solution quality rather than predictive accuracy, by integrating the one-shot optimization problem~\cite{donti2017task,perrault2020end} or sequential problems~\cite{wang2021learning,futoma2020popcorn} as a differentiable layer in the training pipeline. Unfortunately, while
decision-focused learning can successfully optimize the evaluation objective, it is computationally extremely expensive due to the presence of the optimization problems in the training process.
Specifically, for RMAB problems, the computation cost of decision-focused learning arises from the complexity of the sequential problems formulated as Markov decision processes (MDPs), which limits the applicability to RMAB problems due to the PSPACE hardness of finding the optimal solution~\cite{papadimitriou1994complexity}.

Our main contribution is a novel and scalable approach for decision-focused learning in RMAB problems using Whittle index policy, a commonly used approximate solution in RMABs.
Our three key contributions are (i) we establish the differentiability of Whittle index policy to support decision-focused learning to directly optimize the RMAB solution quality; (ii) we show that our approach of differentiating through Whittle index policy improves the scalability of decision-focused learning in RMAB; (iii) we apply our algorithm to an anonymized maternal and child health RMAB dataset previously collected by ~\citet{armman} to evaluate the performance of our algorithm in simulation.

We establish the differentiability of Whittle index by showing that Whittle index can be expressed as a solution to a full-rank linear system reduced from Bellman equations with transition dynamics as entries, which allows us to compute the derivative of Whittle index with respect to transition dynamics. On the other hand, to execute Whittle index policy, the standard selection process of choosing arms with top-k Whittle indices to pull is non-differentiable. We relax this non-differentiable process by using a differentiable soft top-k selection to establish differentiability. 
Our differentiable Whittle index policy enables decision-focused learning in RMAB problems to backpropagate from final policy performance to the predictive model. 
We significantly improve the scalability of decision-focused learning, where the computation cost of our algorithm $O(NM^{\omega+1})$ scales linearly in the number of arms $N$ and polynomially in the number of states $M$ with $\omega \approx 2.373$, while previous work scales exponentially $O(M^{\omega N})$. This significant reduction in computation cost is crucial for extending decision-focused learning to RMAB problems with large number of arms.

In our experiments, we apply decision-focused learning to RMAB problems to optimize importance sampling-based evaluation on synthetic datasets as well as an anonymized RMAB dataset about a maternal and child health program previously collected by~\cite{armman} -- these datasets are the basis of comparing different methods in simulation. We compare decision-focused learning with the two-stage method that trains to minimize the predictive loss.
The two-stage method achieves the best predictive loss but significantly degraded solution quality. In contrast, decision-focused learning reaches a slightly worse predictive loss but with a much better importance sampling-based solution quality evaluation and the improvement generalizes to the simulation-based evaluation that is built from the data.
Lastly, the scalability improvement is the crux of applying decision-focused learning to real-world RMAB problems: our algorithm can run decision-focused learning on the maternal and child health dataset with hundreds of arms, whereas state of the art is a 100-fold slower even with 20 arms and grows exponentially worse.

\section*{Related Work}
\paragraph{Restless multi-armed bandits with given transition dynamics}
This line of research primarily focuses on solving RMAB problems to get a sequential policy. The complexity of solving RMAB problems optimally is known to be PSPACE hard~\cite{papadimitriou1994complexity}. One approximate solution is proposed by~\citet{whittle1988restless}, where they use Lagrangian relaxation to decompose arms and compute the associated Whittle indices to define a policy. Specifically, the indexability condition~\cite{akbarzadeh2019restless,wang2019opportunistic} guarantees this Whittle index policy to be asymptotically optimal~\cite{weber1990index}.
In practice, Whittle index policy usually provides a near-optimal solution to RMAB problems.

\paragraph{Restless multi-armed bandits with missing transition dynamics}
When the transition dynamics are unknown in RMAB problems but an interactive environment is available, prior works~\cite{tekin2012online,liu2012learning,oksanen2015order,dai2011non} consider this as an online learning problem that aims to maximize the expected reward.
However, these approaches become infeasible when interacting with the environment is expensive, e.g., healthcare problems~\cite{mate2022field}.
In this work, we consider the offline RMAB problem, and each arm comes with an arm feature that is correlated to the transition dynamics and can be learned from the past data.

\paragraph{Decision-focused learning}
The predict-then-optimize framework~\cite{elmachtoub2021smart} is composed of a predictive problem that makes predictions on the parameters of the later optimization problem, and an optimization problem that uses the predicted parameters to come up with a solution, where the overall objective is the solution quality of the proposed solution.
Standard two-stage learning method solves the predictive and optimization problems separately, leading to a mismatch of the predictive loss and the evaluation metric~\cite{huang2019addressing,lambert2020objective,johnson2019survey}.
In contrast, decision-focused learning~\cite{wilder2019melding,mandi2020smart,elmachtoub2020decision} learns the predictive model to directly optimize the solution quality by integrating the optimization problem as a differentiable layer~\cite{amos2017optnet,agrawal2019differentiable} in the training pipeline.
Our offline RMAB problem is a predict-then-optimize problem, where we first (offline) learn a mapping from arm features to transition dynamics from the historical data~\cite{mate2022field,sun2018cell}, and the RMAB problem is solved using the predicted transition dynamics accordingly.
Prior work~\cite{mate2022field} is limited to using two-stage learning to solve the offline RMAB problems. While decision-focused learning in sequential problems were primarily studied in the context of MDPs~\cite{wang2021learning,futoma2020popcorn} they come with an expensive computation cost that immediately becomes infeasible in large RMAB problems.
\section{Model: Restless Multi-armed Bandit}
An instance of the restless multi-armed bandit (RMAB) problem is composed of a set of $N$ arms, each is modeled as an independent Markov decision process (MDP). The $i$-th arm in a RMAB problem is defined by a tuple $(\State, \Action, R_i, P_i)$. $\State$ and $\Action$ are the identical state and action spaces across all arms. $R_i, P_i: \State \times \Action \times \State \rightarrow \R$ are the reward and transition functions associated to arm~$i$. We consider finite state space with $|\State| = \nstates$ fully observable states and action set $\Action = \{ 0,1\}$ corresponding to not pulling or pulling the arm, respectively.
For each arm $i$, the reward is denoted by $R_i(\state_i,\action_i,\state'_i) = R(\state_i)$, i.e., the reward $R(\state_i)$ only depends on the current state $\state_i$, where $R: \State \rightarrow \R$ is a vector of size $\nstates$.
Given the state $\state_i$ and action $\action_i$, $P_i(\state_i, \action_i) = [P_i(\state_i, \action_i, \state'_i) ]_{\state'_i \in \State}$ defines the probability distribution of transitioning to all possible next states $\state'_i \in \State$.

In a RMAB problem, at each time step $t \in [T]$, the learner observes $\boldsymbol\state_t = [ \state_{t,i} ]_{i \in [N]} \in \State^N$, the states of all arms. The learner then chooses action ${\boldsymbol \action}_t = [\action_{t,i}]_{i \in [N]} \in \Action^N$ denoting the pulling actions of all arms, which has to satisfy a budget constraint $\sum\nolimits_{i \in [N]} \action_{t,i} \leq K$, i.e., the learner can pull at most $K$ arms at each time step.
Once the action is chosen, arms receive action $\boldsymbol\action_{t}$ and transitions under $P$ with rewards $\boldsymbol\reward_t = [ \reward_{t,i} ]_{i \in [N]}$ accordingly.
We denote a full trajectory by $\trajectory = (\boldsymbol\state_1, \boldsymbol\action_1, \boldsymbol\reward_1, \cdots, \boldsymbol\state_T, \boldsymbol\action_T, \boldsymbol\reward_T)$.
The total reward is defined by the summation of the discounted reward across $T$ time steps and $N$ arms, i.e., $\sum\nolimits_{t = 1}^T \gamma^{t-1} \sum\nolimits_{i \in [N]} \reward_{t,i}$, where $0 < \gamma \leq 1$ is the discount factor.

A policy is denoted by $\policy$, where $\policy(\boldsymbol\action \mid \boldsymbol\state)$ is the probability of choosing action $\boldsymbol\action$ given state $\boldsymbol\state$. Additionally, we define $\policy(\action_i = 1 \mid \boldsymbol\state)$ to be the marginal probability of pulling arm $i$ given state $\boldsymbol\state$, where $\policy(\boldsymbol\state) = [\policy(\action_i = 1 \mid \boldsymbol\state)]_{i \in [N]}$ is a vector of arm pulling probabilities. Specifically, we use $\policy^*$ to denote the optimal policy that optimizes the cumulative reward, while $\policy^\text{solver}$ to denote a near-optimal policy solver.


\section{Problem Statement}
This paper studies the RMAB problem where we do not know the transition probabilities $P = \{ P_i \}_{i \in [N]}$ in advance.
Instead, we are given a set of features $\boldsymbol\feature = \{\feature_i \in \Feature\}_{i \in [N]}$, each corresponding to one arm. The goal is to learn a mapping $\model_\weight: \Feature \rightarrow \mathcal{P}$, parameterized by weights $\weight$, to make predictions on the transition probabilities $P = \model_\weight(\boldsymbol\feature) \coloneqq \{ \model_\weight(\feature_i) \}_{i \in [N]}$. The predicted transition probabilities are later used to solve the RMAB problem to derive a policy $\policy = \policy^\text{solver}(\model_\weight(\boldsymbol\feature))$. The performance of the model $\model$ is evaluated by the performance of the proposed policy $\policy$.

\subsection{Training and Testing Datasets}
To learn the mapping $\model_\weight$, we are given a set of RMAB instances as training examples $\dataset_{\text{train}} = \{(\boldsymbol\feature, \trajectories) \}$, where each instance is composed of a RMAB problem with feature $\boldsymbol\feature$ that is correlated to the unknown transition probabilities $P$, and a set of realized trajectories $\trajectories = \{\trajectory^{(j)}\}_{j \in J}$ generated from a given behavior policy $\policy_{\text{beh}}$ that determined how to pull arms in the past.
The testing set $\dataset_\text{test}$ is defined similarly but hidden at training time.

\subsection{Evaluation Metrics}
\paragraph{Predictive loss}
To measure the correctness of transition probabilities $P = \{ P_i \}_{i \in [N]}$, we define the predictive loss as the average negative log-likelihood of seeing the given trajectories $\trajectories$, i.e., $\mathcal{L}(P, \trajectories) \coloneqq - \log \Pr(\trajectories \mid P) =  - \mathop{\mathbb{E}}\nolimits_{\trajectory \sim \trajectories} \sum\nolimits_{t \in [T]} \log P(\boldsymbol\state_{t}, \boldsymbol\action_{t}, \boldsymbol\state_{t+1})$. Therefore, we can define the predictive loss of a model $\model_\weight$ on dataset $\dataset$ by:
\begin{align}\label{eqn:predictive-loss}
    \mathop{\mathbb{E}}\nolimits_{(\boldsymbol\feature, \trajectories) \sim \dataset} \mathcal{L}(\model_\weight(\boldsymbol\feature), \trajectories)
\end{align}

\paragraph{Policy evaluation}
On the other hand, given transition probabilities $P$, we can solve the RMAB problem to derive a policy $\policy^\text{solver}(P)$. 
We can use the historical trajectories $\trajectories$ to evaluate how good the policy performs, denoted by $\text{Eval}(\policy^\text{solver}(P), \trajectories)$. Given dataset $\dataset$, we can evaluate the predictive model $\model_\weight$ on dataset $\dataset$ by:
\begin{align}\label{eqn:policy-evaluation}
    \mathop{\mathbb{E}}\nolimits_{(\boldsymbol\feature, \trajectories) \sim \dataset} \text{Eval}(\policy^\text{solver}(\model_\weight(\boldsymbol\feature)), \trajectories)
\end{align}
Two common types of policy evaluation are importance sampling-based off-policy policy evaluation and simulation-based evaluation, which will be discussed in Section~\ref{sec:policy-evaluation}.


\subsection{Learning Methods}
\begin{figure*}
    \centering
    \includegraphics[width=0.95\linewidth]{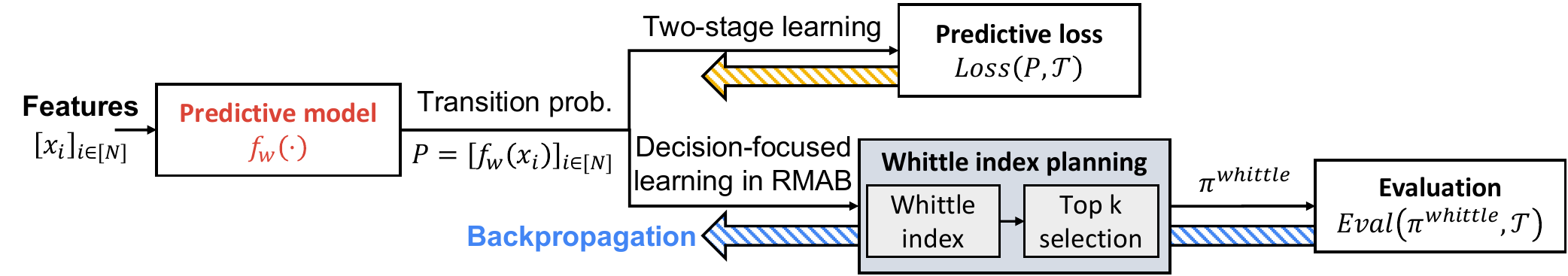}
    \caption{This flowchart visualizes different methods of learning the predictive model. Two-stage learning directly compares the predicted transition probabilities with the given data to define a predictive loss to run gradient descent. Decision-focused learning instead goes through a policy solver using Whittle index policy to estimate the final evaluation and run gradient ascent.}
    \label{fig:flowchart}
\end{figure*}

\paragraph{Two-stage learning} To learn the predictive model $\model_\weight$, we can minimize Equation~\ref{eqn:predictive-loss} by computing gradient $\frac{d \mathcal{L}( \model_\weight(\boldsymbol\feature), \trajectories)}{d \weight}$ to run gradient descent. However, this training objective (Equation~\ref{eqn:predictive-loss}) differs from the evaluation objective (Equation~\ref{eqn:policy-evaluation}), which often leads to suboptimal performance.

\paragraph{Decision-focused learning} In contrast, we can directly run gradient ascent to maximize Equation~\ref{eqn:policy-evaluation} by computing the gradient $\frac{d \text{Eval}(\policy^{\text{solver}}(\model_\weight(\boldsymbol\feature)), \trajectories)}{d \weight}$. However, in order to compute the gradient, we need to differentiate through the policy solver $\policy^{\text{solver}}$ and the corresponding optimal solution.
Unfortunately, finding the optimal policy in RMABs is expensive and the policy is high-dimensional.
Both of these challenges prevent us from computing the gradient to achieve decision-focused learning.





\section{Decision-focused Learning in RMABs}\label{sec:approximate-decision-focused}
In this paper, instead of grappling with the optimal policy, we consider the Whittle index policy~\cite{whittle1988restless} -- the dominant solution paradigm used to solve the RMAB problem.
Whittle index policy is easier to compute and has been shown to perform well in practice. In this section we establish that it is also possible to backpropagate through the Whittle index policy.
This differentiability of Whittle index policy allows us to run decision-focused learning to directly maximize the performance in the RMAB problem.

\subsection{Whittle Index and Whittle Index Policy}\label{sec:whittle-index-policy}

Informally, the Whittle index of an arm captures the added value derived from pulling that arm. The key idea is to determine the Whittle indices of all arms and to pull the arms with the highest values of the index.  

To evaluate the value of pulling an arm $i$, we consider the notion of `passive subsidy', which is a hypothetical exogenous compensation $\whittle$ rewarded for not pulling the arm (i.e. for choosing action $a=0$). Whittle index is defined as the smallest subsidy necessary to make pulling as rewarding as not pulling, assuming indexability~\cite{liu2010indexability}:
\begin{definition}[Whittle index]\label{def:whittle-index}
Given state $\tmpstate \in \State$, we define the Whittle index associated to state $\tmpstate$ by:
\begin{align}
    W_i(\tmpstate) & \coloneqq \inf\nolimits_{\whittle}
    \{ Q_{i}^\whittle(\tmpstate; a=0) = Q_{i}^\whittle(\tmpstate; a=1)\} \label{eqn:action-indifference}
\end{align}
where the value functions are defined by the following Bellman equations, augmented with subsidy $\whittle$ for action $a=0$.
\begin{align}
    V^{\whittle}_i(s) &=  \max\nolimits_{\action} Q_{i}^\whittle(\state; \action)  \label{eqn:bellman-equation} \\
    Q_{i}^\whittle(\state; \action) \! &= \! \whittle \boldsymbol 1_{a = 0} \! + \! R(\state) \! + \! \gamma \! \sum\nolimits_{\state'} \! \parameter_{i}(\state,a,\state') V^{\whittle}_{i}(s') \! \label{eqn:bellman-equation2}
\end{align}
\end{definition}

Given the Whittle indices of all arms and all states $W = [W_i(\tmpstate)]_{i \in [N], \tmpstate \in \State}$, the Whittle index policy is denoted by $\policy^\text{whittle}: \State^N \longrightarrow [0,1]^N$, which takes the states of all arms as input to compute their Whittle indices and output the probabilities of pulling arms. This policy repeats for every time step to pull arms based on the index values.

\subsection{Decision-focused Learning Using Whittle Index Policy}
Instead of using the optimal policy $\policy^*$ to run decision-focused learning with expensive computation cost, we use Whittle index policy $\policy^{\text{whittle}}$ to determine how to pull arms as an approximate solution. In this case, in order to run decision-focused learning, we need to compute the derivative of the evaluation metric by chain rule:
\begin{align}\label{eqn:differentiable-whittle-policy-derivative}
     \frac{d \text{Eval}(\policy^\text{whittle}, \trajectories)}{d \weight} = \frac{d \text{Eval}(\policy^\text{whittle}, \trajectories)}{d \policy^\text{whittle}} \frac{d \policy^\text{whittle}}{d W} \frac{d W}{d P} \frac{d P}{d \weight}
\end{align}
where $W$ is the Whittle indices of all states under the predicted transition probabilities $P$. The policy $\policy^{\text{whittle}}$ is the Whittle index policy induced by $W$. The flowchart is illustrated in Figure~\ref{fig:flowchart}.

The term $\frac{d \text{Eval}(\policy^\text{whittle}, \trajectories)}{d \policy^\text{whittle}}$ can be computed via policy gradient theorem~\cite{sutton1998introduction}, and the term $\frac{d P}{d \weight}$ can be computed using auto-differentiation.
However, there are still two challenges remaining: (i) how to differentiate through Whittle index policy to get $\frac{d \policy^\text{whittle}}{d W}$ (ii) how to differentiate through Whittle index computation to derive $\frac{d W}{d P}$. 

\subsection{Differentiability of Whittle Index Policy}
A common choice of Whittle index policy is defined by:
\begin{definition}[Strict Whittle index policy]
\begin{align}
    \policy^{\text{strict}}_W(\boldsymbol\state) = \boldsymbol1_{\text{top-k}([W_i(\state_i)]_{i \in [N]})} \in \{0,1\}^N
\end{align}
which selects arms with the top-k Whittle indices to pull.
\end{definition}

However, the strict top-k operation in the strict Whittle index policy is non-differentiable, which prevents us from computing a meaningful estimate of $\frac{d \policy^\text{whittle}}{d W}$ in Equation~\ref{eqn:differentiable-whittle-policy-derivative}.
We circumvent this issue by relaxing the top-k selection to a soft-top-k selection~\cite{xie2020differentiable}, which can be expressed as an optimal transport problem with regularization, making it differentiable.
We apply soft-top-k to define a new differentiable soft Whittle index policy:
\begin{definition}[Soft Whittle index policy]
\begin{align}\label{eqn:differentiable-whittle-policy}
    \policy^{\text{soft}}_W(\boldsymbol\state) = \text{soft-top-k}([W_j(\state_i)]_{i \in [N]}) \in [0,1]^N
\end{align}
\end{definition}
Using the soft Whittle index policy, the policy becomes differentiable and we can compute $\frac{d \policy^{\text{whittle}}}{d W}$.

\subsection{Differentiability of Whittle Index}
The second challenge is the differentiability of Whittle index. Whittle indices are often computed using value iteration and binary search~\cite{qian2016restless,mate2020collapsing} or mixed integer linear program. However, these operations are not differentiable and we cannot compute the derivative $\frac{d W}{d P}$ in Equation~\ref{eqn:differentiable-whittle-policy-derivative} directly.


\begin{figure*}
    \centering
    \includegraphics[width=0.98\linewidth]{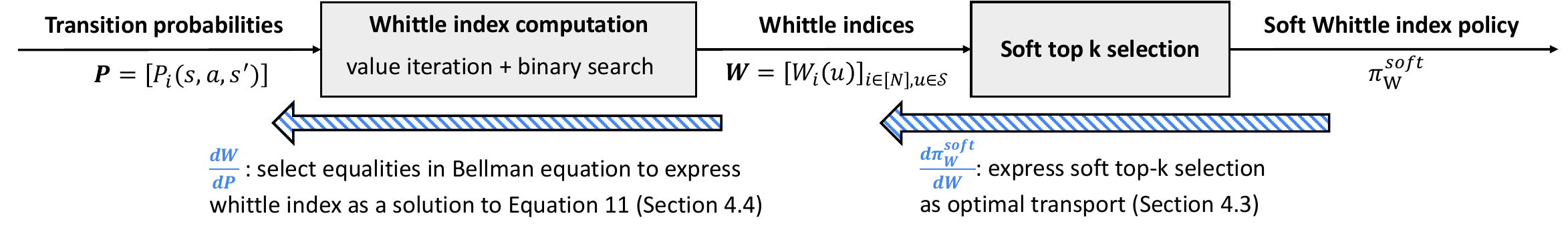}
    \caption{We establish the differentiability of Whittle index policy using a soft top-k selection to construct a soft Whittle index policy, and the differentiability of Whittle index by expressing Whittle index as a solution to a linear system in Equation~\ref{eqn:selected-bellman-equation}.}
    \label{fig:whittle}
\end{figure*}


\paragraph{Main idea}
After computing the Whittle indices and the value functions of each arm $i$, the key idea is to construct linear equations that link the Whittle index with the transition matrix $P_i$. Specifically, we achieve this by resolving the $\max$ operator in Equation~\ref{eqn:bellman-equation} of Definition~\ref{def:whittle-index} by determining the optimal actions $a$ from the pre-computed value functions. Plugging back in Equation~\ref{eqn:bellman-equation2} and manipulating as shown below yields linear equations in the Whittle index $W_i(u)$ and transition matrix $P_i$, which can be expressed as a full-rank linear system in $P_i$, with the Whittle index as a solution. This makes the Whittle index differentiable in $P_i$.

\paragraph{Selecting Bellman equation}
Let $\tmpstate$ and arm $i$ be the target state and target arm to compute the Whittle index.
Assume we have precomputed the Whittle index $m = W_i(\tmpstate)$ for state $u$ and the corresponding value functions $[V^\whittle_{i}(\state)]_{\state \in \State}$ for all states under the same passive subsidy $m = W_i(\tmpstate)$.
Equation~\ref{eqn:bellman-equation2} can be combined with Equation~\ref{eqn:bellman-equation} to get:
\begin{align}
    V^{\whittle}_{i}(s) \geq \begin{cases}
    \whittle + R(\state) + \gamma \sum\nolimits_{\state' \in \State} \parameter_{i}(\state,a=0,\state') V^{\whittle}_{i}(s') \\
    R(\state) + \gamma \sum\nolimits_{\state' \in \State} \parameter_{i}(\state,a=1,\state') V^{\whittle}_{i}(s')
    \end{cases} \label{eqn:linear-equation}
\end{align}
where $m= W_i(\tmpstate)$.

For each $\state \in S$, at least one of the equalities in Equation~\ref{eqn:linear-equation} holds because one of the actions must be optimal and match the state value function $V^\whittle_{i}(\state)$. We can identify which equality holds by simply plugging in values of precomputed value functions $[V^\whittle_{i}(\state)]_{\state \in \State}$. Furthermore, for the target state $\tmpstate$, both equalities must hold because by the definition of Whittle index, the passive subsidy $m= W_i(\tmpstate)$ makes both actions equally optimal, i.e.  
in Equation~\ref{eqn:action-indifference}, $V^{\whittle}_{i}(\tmpstate) = Q^{\whittle}_{i}(\tmpstate,\action=0) = Q^{\whittle}_{i}(\tmpstate,\action=1)$ for $\whittle = W_i(\tmpstate)$.

Thus Equation~\ref{eqn:linear-equation} can be written in matrix form:
\begin{align}
\begin{bmatrix}
    {\boldsymbol V}^\whittle_i \\
    {\boldsymbol V}^\whittle_i
\end{bmatrix}
\! \geq \!
\begin{bmatrix}
   \boldsymbol 1_{\nstates} \! & \! \gamma \boldsymbol P_i(\State, a\!=\!0, \State) \\
   \boldsymbol 0_{\nstates} \! & \! \gamma \boldsymbol P_i(\State, a\!=\!1, \State)
\end{bmatrix}
\begin{bmatrix}
    \whittle \\
    {\boldsymbol V}^\whittle_i
\end{bmatrix}
\! + \!
\begin{bmatrix}
\boldsymbol R(S) \\
\boldsymbol R(S)
\end{bmatrix}
\label{eqn:linear-equation-matrix-form}
\end{align}
where $\boldsymbol V^\whittle_i \coloneqq [V^\whittle_i(\state)]_{\state \in \State}$, $\boldsymbol R(\State) = [R(\state)]_{\state \in \State}$, and $\boldsymbol P_i(\State, \action,\State) \coloneqq [P_i(\state, \action, \state')]_{\state, \state' \in \State} \in \R^{\nstates \times \nstates}$.

By the aforementioned discussion, we know that there are at least $M+1$ equalities in Equation~\ref{eqn:linear-equation-matrix-form} while there are also only $M+1$ variables ($m \in \R$ and $\boldsymbol V^\whittle_i \in \R^M$).
Therefore, we rearrange Equation~\ref{eqn:linear-equation-matrix-form} and pick only the rows where equalities hold to get:
\begin{align}\label{eqn:selected-bellman-equation}
A
\begin{bmatrix}
   \boldsymbol 1_{\nstates} & \gamma \boldsymbol P_i(\State, a=0, \State) - I_{\nstates} \\
   \boldsymbol 0_{\nstates} & \gamma \boldsymbol P_i(\State, a=1, \State) - I_{\nstates}
\end{bmatrix}
\begin{bmatrix}
    \whittle \\
    {\boldsymbol V}^\whittle_i
\end{bmatrix}
=
A
\begin{bmatrix}
- \boldsymbol R(S) \\
- \boldsymbol R(S)
\end{bmatrix}
\end{align}
where we use a binary matrix $A \in \{0,1\}^{(M+1) \times 2M}$ with a single $1$ per row to extract the equality. For example, we can set $A_{ij} = 1$ if the $j$-th row in Equation~\ref{eqn:linear-equation-matrix-form} corresponds to the equality in Equation~\ref{eqn:linear-equation} with the $i$-th state in the state space $S$ for $i \in [M]$, and the last row $A_{(M+1),j} = 1$ to mark the additional equality matched by the Whittle index definition (see Appendix~\ref{sec:differentiability-details} for more details). Matrix $A$ picks $M+1$ equalities out from Equation~\ref{eqn:linear-equation-matrix-form} to form Equation~\ref{eqn:selected-bellman-equation}.


Equation~\ref{eqn:selected-bellman-equation} is a full-rank linear system with $m = W_i(\tmpstate)$ as a solution. 
This expresses $W_i(\tmpstate)$ as an implicit function of $\boldsymbol P$, allowing for computation of $\frac{d W_i(\tmpstate)}{d \boldsymbol P}$ via autodifferentiation, thus achieving differentiability of the Whittle index.
We repeat this process for every arm $i \in [N]$ and every state $\tmpstate$.
Figure~\ref{fig:whittle} summarizes the differentiable Whittle index policy and the algorithm is shown in Algorithm~\ref{alg:approximate-decision-focused-learning}.

\subsection{Computation Cost and Backpropagation}\label{sec:computation-cost}
It is well studied that Whittle index policy can be computed more efficiently than solving the RMAB problem as a large MDP problem.
Here, we show that the use of Whittle index policy also demonstrates a large speed up in terms of backpropagating the gradient in decision-focused learning.

In order to use Equation~\ref{eqn:selected-bellman-equation} to compute the gradient of Whittle indices, we need to invert the left-hand-side of Equation~\ref{eqn:selected-bellman-equation} with dimensionality $M+1$, which takes $O(M^\omega)$ where $\omega \approx 2.373$~\cite{alman2021refined} is the best known matrix inversion constant.
Therefore, the overall computation of all $N$ arms and $M$ states is $O(N M^{\omega + 1})$ per gradient step.

In contrast, the standard decision-focused learning differentiates through the optimal policy using the full Bellman equation with $O(M^N)$ variables, where inverting the large Bellman equation requires $O(M^{\omega N})$ cost per gradient step.
Thus, our algorithm significantly reduces the computation cost to a linear dependency on the number of arms $N$. This significantly improves the scalability of decision-focused learning. 

\subsection{Extension to Partially Observable RMAB}\label{sec:partially-observable}
For partially observable RMAB problem, we focus on a subclass of RMAB problem known as collapsing bandits~\cite{mate2020collapsing}. In collapsing bandits,
belief states~\cite{monahan1982state} are used to represent the posterior belief of the unobservable states.
Specifically, for each arm $i$, we use $\belief_i \in \Belief = \Delta(\State) \subset [0,1]^M$ to denote the posterior belief of an arm, where each entry $\belief_i(\state_i)$ denotes the probability that the true state is $\state_i \in \State$. When arm $i$ is pulled, the current true state $\state_i \sim \belief_i$ is revealed and drawn from the posterior belief with expected reward $\belief_i^\top R$, where we can define the transition probability on the belief states. This process reduces partially observable states to fully observable belief states with in total $MT$ states since the maximal horizon is $T$. Therefore, we can use the same technique to differentiate through Whittle indices of partially observable states.


\section{Policy Evaluation Metrics}\label{sec:policy-evaluation}
In this paper, we use two different variants of evaluation metric: importance sampling-based evaluation~\cite{sutton1998introduction} and simulation-based (model-based) evaluation.



\paragraph{Importance sampling-based Evaluation}
We adopt Consistent Weighted Per-Decision Importance Sampling (CWPDIS)~\cite{thomas2015safe} as our importance sampling-based evaluation.
Given target policy $\policy$ and a trajectory $\trajectory = \{ \state_1, \action_1, \reward_1, \cdots, \state_T, \action_T, \reward_T \}$ executed by the behavior policy $\policy_{\text{beh}}$, the importance sampling weight is defined by $\rho_{ti} = \prod\nolimits_{t'=1}^t \frac{\policy(\action_{t',i} \mid \state_{t'})}{\policy_{\text{beh}}(\action_{t',i} \mid \state_{t'})}$. We evaluate the policy $\policy$ by: 
\begin{align}\label{eqn:IS-OPE}
    \text{Eval}_\text{IS}(\policy, \trajectories) = \sum\nolimits_{t \in [T], i \in [N]} \gamma^{t-1} \frac{\mathop{\mathbb{E}}_{\trajectory \sim \trajectories} \left[ \reward_{t,i} \rho_{ti}(\trajectory) \right]}{\mathop{\mathbb{E}}_{\trajectory \sim \trajectories} \left[ \rho_{ti}(\trajectory) \right]}
\end{align}
Importance sampling-based evaluations are often unbiased but with a larger variance due to the unstable importance sampling weights. CWPDIS normalizes the importance sampling weights to achieve a consistent estimate. 



\paragraph{Simulation-based Evaluation}
An alternative way is to use the given trajectories to construct an empirical transition probability $\bar{P}$ to build a simulator and evaluate the target policy $\policy$.
The variance of simulation-based evaluation is small, but it may require additional assumptions on the missing transition when the empirical transition $\bar{P}$ is not fully reconstructed.

\begin{algorithm}[tb]
   \caption{Decision-focused Learning in RMAB}
   \label{alg:approximate-decision-focused-learning}
\begin{algorithmic}[1]
   \STATE {\bfseries Input:} training set $\dataset_\text{train}$, learning rate $r$, model $\model_\weight$
   \FOR{epoch $= 1,2,\cdots$ and $(\feature, \trajectories) \in \dataset_\text{train}$}
   \STATE Predict $P = \model_\weight(\feature)$ and compute Whittle indices $W(P)$. 
   \STATE Let $\policy^\text{whittle} = \policy^{\text{soft}}_W$ and compute $\text{Eval}(\policy^\text{whittle}, \trajectories)$.
   \STATE Update $\weight = \weight + r \frac{d \text{Eval}(\policy^\text{whittle}, \trajectories)}{d \policy^\text{whittle}} \frac{d \policy^\text{whittle}}{d W} \frac{d W}{d P} \frac{d P}{d \weight}$, where $\frac{d W}{d P}$ is computed from Equation~\ref{eqn:selected-bellman-equation}.
   \ENDFOR
   \STATE {\bfseries Return:} predictive model $\model_\weight$
\end{algorithmic}
\end{algorithm}

\section{Experiments}\label{sec:experiments}

\begin{figure*}[t]
    \centering
    \begin{subfigure}{\linewidth}
        \centering
        \includegraphics[width=0.4\textwidth]{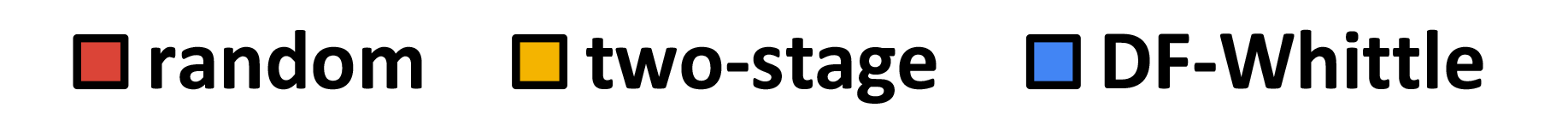}
    \end{subfigure}
    \centering
    \begin{subfigure}{0.325\linewidth}
        \centering
        \includegraphics[width=\textwidth]{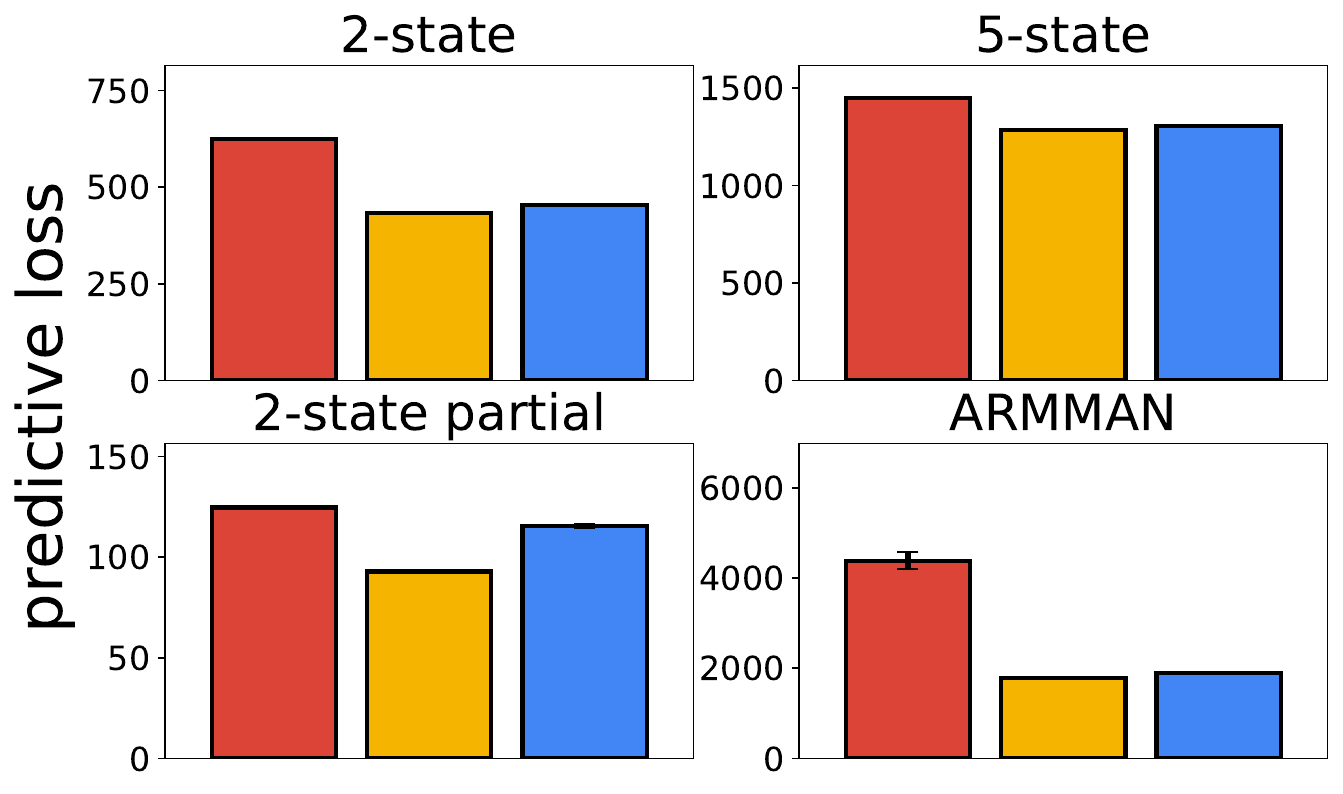}
        \caption{Predictive loss}
        \label{fig:barchart-loss}
    \end{subfigure}
    \hfill
    \begin{subfigure}{0.325\linewidth}
        \centering
        \includegraphics[width=\textwidth]{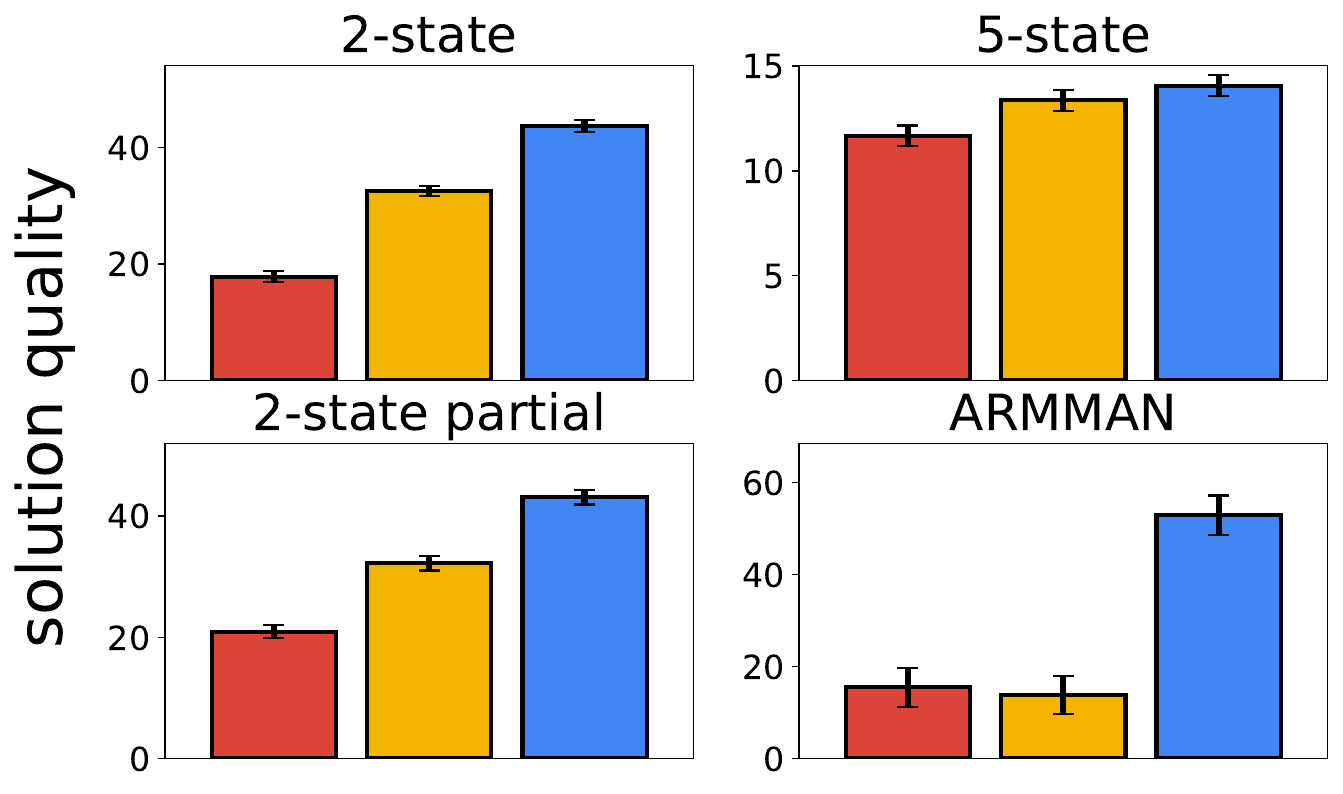}
        \caption{IS-based evaluation}
        \label{fig:barchart-is}
    \end{subfigure}
    \hfill
    \begin{subfigure}{0.325\linewidth}
        \centering
        \includegraphics[width=\textwidth]{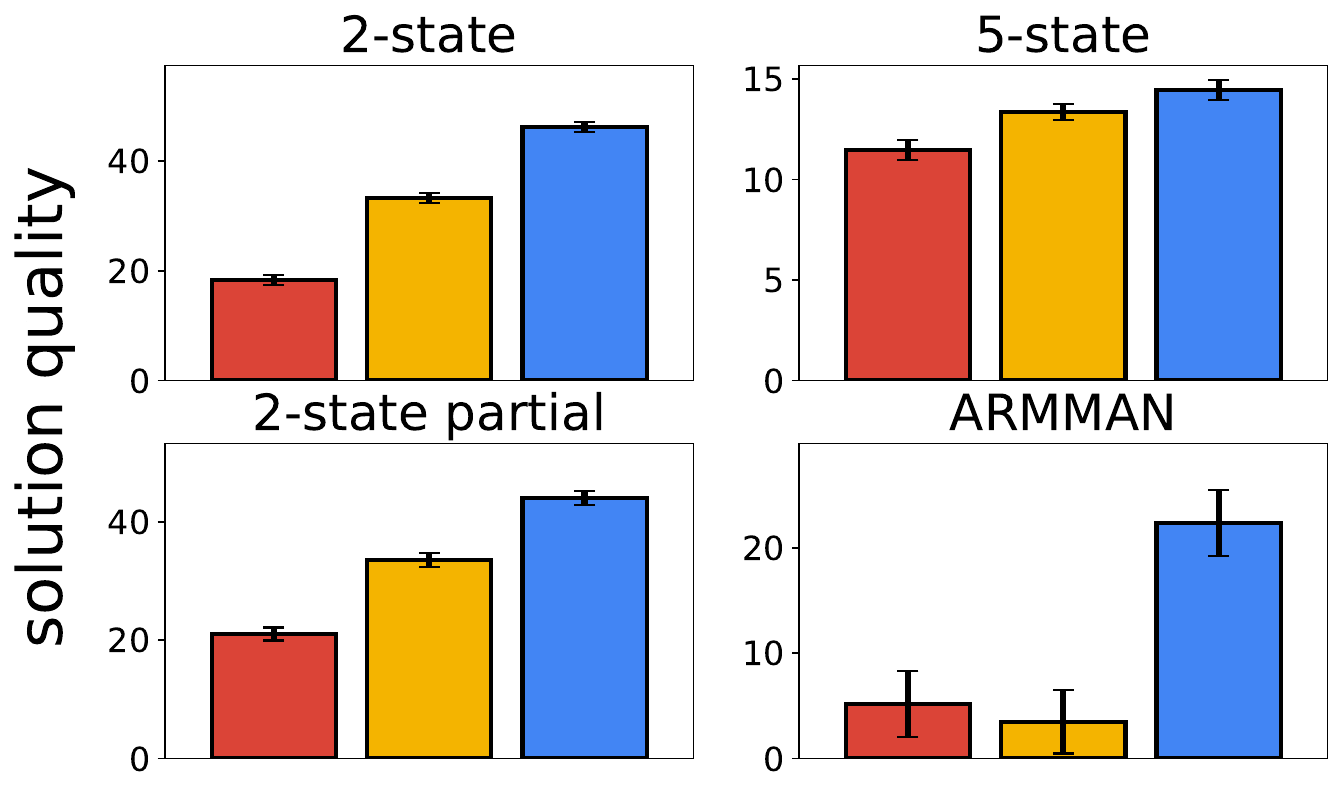}
        \caption{Simulation-based evaluation}
        \label{fig:barchart-sim}
    \end{subfigure}
    \caption{Comparison of predictive loss, importance sampling-based evaluation, and simulation-based evaluation on all synthetic domains and the real ARMMAN dataset. 
    For the evaluation metrics, we plot the improvement against the no-action baseline that does not pull any arm.
    Although two-stage method achieves the smallest predictive loss, decision-focused learning consistently outperforms two-stage method in both \textit{solution quality} evaluation metrics across all domains. 
    }
    \label{fig:barchart}
\end{figure*}

We compare two-stage learning ({\bf TS}) with our decision-focused learning ({\bf DF-Whittle}) that optimizes importance sampling-based evaluation directly.
We consider three different evaluation metrics including predictive loss, importance sampling evaluation, and simulation-based evaluation to evaluate all learning methods.
We perform experiments on three synthetic datasets including $2$-state fully observable, $5$-state fully observable, and $2$-state partially observable RMAB problems. We also perform experiments on a real dataset on maternal and child health problem modelled as a $2$-state fully observable RMAB problem with real features and historical trajectories. For each dataset, we use $70\%$, $10\%$, $20\%$ of the RMAB problems as the training, validation, and testing sets, respectively. All experiments are averaged over $50$ independent runs.

\paragraph{Synthetic datasets}
We consider RMAB problems composed of $N = 100$ arms, $\nstates$ states, budget $K=20$, and time horizon $T=10$ with a discount rate of $\gamma = 0.99$.
The reward function is given by $R = [\frac{i-1}{\nstates-1}]_{i \in [\nstates]}$, while the transition probabilities are generated uniformly at random but with a constraint that pulling the arm ($a=1$) is strictly better than not pulling the arm ($a=0$) to ensure the benefit of pulling.
To generate the arm features, we feed the transition probability of each arm to a randomly initialized neural network to generate fixed-length correlated features with size $16$ per arm. 
The historical trajectories $\trajectories$ with $|\trajectories| = 10$ are produced by running a random behavior policy $\policy_{\text{beh}}$.
The goal is to predict transition probabilities from the arm features and the training trajectories.

\paragraph{Real dataset}
The Maternal and Child Healthcare Mobile Health program operated by~\citet{armman} aims to improve dissemination of health information to pregnant women and mothers with an aim to reduce maternal, neonatal and child mortality and morbidity. ARMMAN serves expectant/new mothers in disadvantaged communities with \textit{median daily family income of \$3.22 per day} which is seen to be below the world bank poverty line~\cite{world2020poverty}. The program is composed of multiple enrolled beneficiaries and a planner who schedules service calls to improve the overall engagement of beneficiaries; engagement is measured in terms of total number of automated voice (health related) messages that the beneficiary engaged with. More precisely, this problem is modelled as a $M=2$-state fully observable RMAB problem where each beneficiary's behavior is governed by an MDP with two states - Engaging and Non-Engaging state; engagement is determined by whether the beneficiary listens to an automated voice message (average length 115 seconds) for more than 30 seconds. The planner's task is to recommend a subset of beneficiaries every week to receive service calls from health workers to further improve their engagement behavior. We do not know the transition dynamics, but we are given beneficiaries' socio-demographic features to predict transition dynamics.

We use a subset of data from the large-scale anonymized quality improvement study performed by ARMMAN for $T=7$ weeks, obtained from~\citet{mate2022field}, with beneficiary consent. In the study, a cohort of beneficiaries received Round-Robin policy, scheduling service calls in a fixed order, with a single trajectory $|\trajectories| = 1$ per beneficiary that documents the calling decisions and the engagement behavior in the past. We randomly split the cohort into $8$ training groups, $1$ validation group, and $3$ testing groups each with $N=639$ beneficiaries and $K=18$ budget formulated as an RMAB problem. The demographic features of beneficiaries are used to infer the missing transition dynamics.

\paragraph{Data usage}
All the datasets are anonymized. The experiments are secondary analysis using different evaluation metrics with approval from the ARMMAN ethics board. There is no actual deployment of the proposed algorithm at ARMMAN.
For more details about the dataset, consent of data collection, please refer to Appendix~\ref{sec:dataset} and~\ref{sec:consent}.

\begin{figure}[t]
    \centering
    \includegraphics[width=0.65\linewidth]{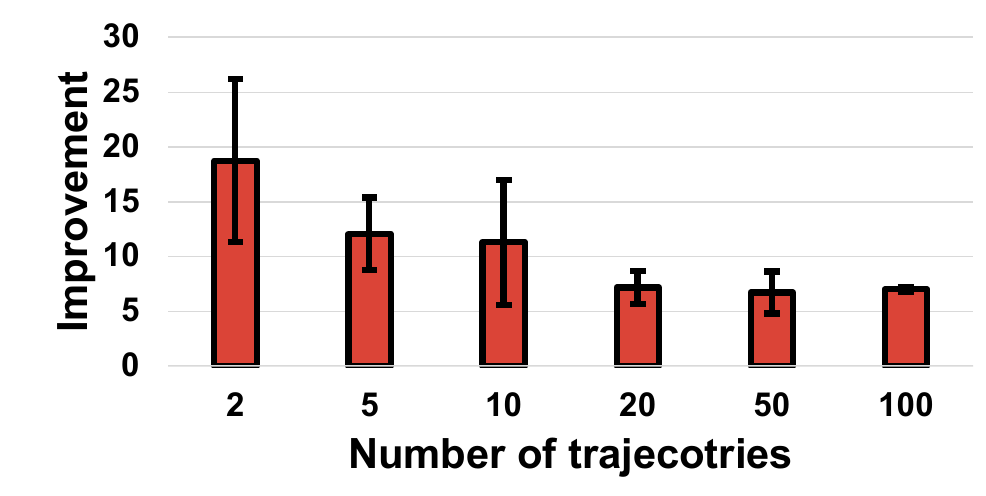}
    \caption{Performance improvement of decision-focused v.s. two-stage method with varying number of trajectories.}
    \label{fig:trajectory}
\end{figure}

\begin{figure}
    \begin{subfigure}{0.47\linewidth}
        \includegraphics[width=\textwidth]{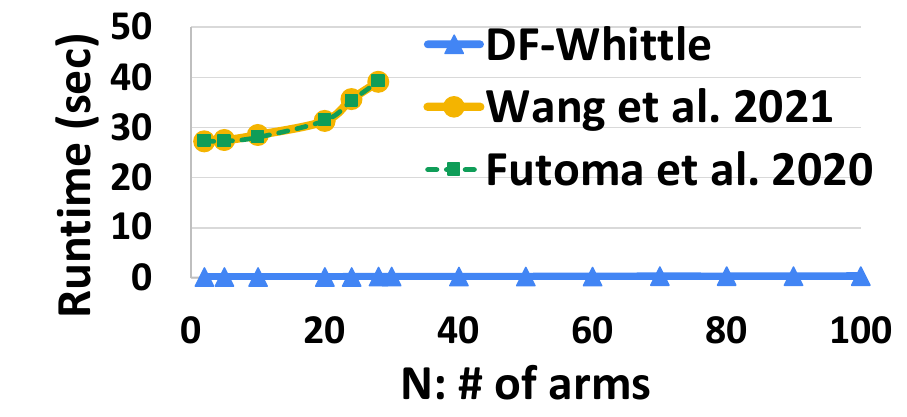}
        \caption{Comparing out algorithm to decision-focused baselines.}
        \label{fig:comp_cost_baselines}
    \end{subfigure}
    \hfill
    \begin{subfigure}{0.47\linewidth}
        \includegraphics[width=\textwidth]{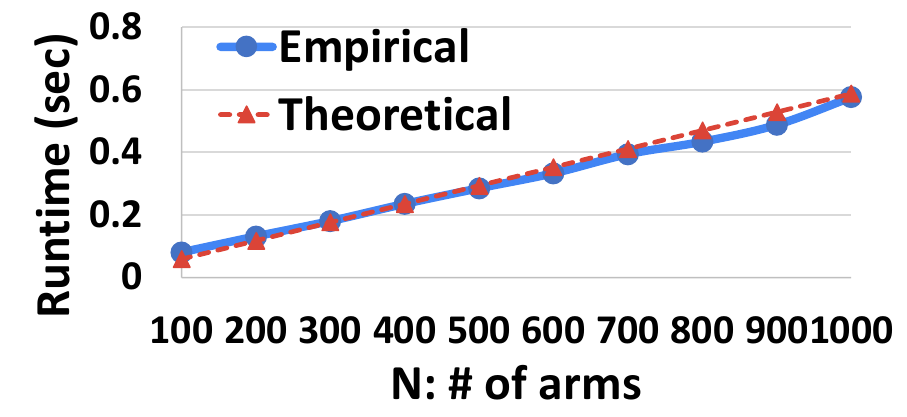}
        \caption{Computation cost with varying number of arms $N$.}
        \label{fig:com_cost_arms}
    \end{subfigure}
    \caption{We compare the computation cost of our decision-focused learning with other baselines and the theoretical complexity $O(NM^{\omega + 1})$ with varying number of arms $N$.}
\end{figure}

\section{Experimental Results}\label{sec:results}
\paragraph{Performance improvement and justification of objective mismatch}
In Figure~\ref{fig:barchart}, we show the performance of random policy, two-stage, and decision-focused learning (DF-Whittle) on three evaluation metrics - predictive loss, importance sampling-based evaluation and simulation-based evaluation for all domains. For the evaluation metrics, we plot the improvement against the no-action baseline that does not pull any arms throughout the entire RMAB problem. We observe that two-stage learning consistently converges to a smaller predictive loss, while DF-Whittle outperforms two-stage on all solution quality evaluation metrics significantly (p-value $<$ $0.05$) by alleviating the objective mismatch issue. This result also provides evidence of aforementioned objective mismatch, where the advantage of two-stage in the predictive loss does not translate to solution quality.

\paragraph{Significance in maternal and child care domain}
In the ARMMAN data in Figure~\ref{fig:barchart}, we assume limited resources that we can only select $18$ out of $638$ beneficiaries to make service call per week. Both random and two-stage method lead to around $15$ more (IS-based evaluation) listening to automated voice messages among all beneficiaries throughout the 7-week program by $18 \times 7 = 126$ service calls, when compared to not scheduling any service call; this low improvement also reflects the hardness of maximizing the effectiveness of service calls. In contrast, decision-focused learning achieves an increase of beneficiaries listening to 50 more voice messages overall; DF-whittle achieves a much higher increase by strategically assigning the limited service calls using the right objective in the learning method. The improvement is statistically significant (p-value $<$ $0.05$).

In the testing set, we examine the difference between those selected for service call in two-stage and DF-Whittle. We observe that there are some interesting differences. For example, DF-Whittle chooses to do service calls to expectant mothers earlier in gestational age (22\% vs 37\%), and to a lower proportion of those who have already given birth (2.8\% vs 13\%) compared to two-stage. In terms of the income level, there is no statistic significance between two-stage and DFL (p-value = 0.20 see Appendix~\ref{sec:dataset}). In particular, 94\% of the mothers selected by both methods are below the poverty line~\cite{world2020poverty}.

\paragraph{Impact of Limited Data}
Figure~\ref{fig:trajectory} shows the improvement between decision-focused learning and two-stage method with varying number of trajectories given to evaluate the impact of limited data. We notice that a larger improvement between decision-focused and two-stage learning is observed when fewer trajectories are available.
We hypothesize that less samples implies larger predictive error and more discrepancy between the loss metric and the evaluation metric.

\paragraph{Computation cost comparison}
Figure \ref{fig:comp_cost_baselines}, compares the computation cost per gradient step of our Whittle index-based decision-focused learning and other baselines in decision-focused learning~\cite{wang2021learning,futoma2020popcorn} by changing $N$ (the number of arms) in $M=2$-state RMAB problem.
The other baselines fail to run with $N=30$ arms and do not scale to larger problems like maternal and child care with more than $600$ people enrolled, while our approach is 100x faster than the baselines as shown in Figure~\ref{fig:comp_cost_baselines} and with a linear dependency on the number of arms $N$.

In Figure \ref{fig:com_cost_arms}, we compare the empirical computation cost of our algorithm with the theoretical computation complexity $O(N M^{\omega +1})$ in $N$ arms and $M$ states RMAB problems.
The empirical computation cost matches with the linear trend in $N$. Our computation cost significantly improves the computation cost $O(M^{\omega N})$ of previous work as discussed in Section~\ref{sec:computation-cost}.

\section{Conclusion}
This paper presents the first decision-focused learning in RMAB problems that is scalable for large real-world datasets. We establish the differentiability of Whittle index policy in RMAB by providing new method to differentiate through Whittle index and using soft-top-k to relax the arm selection process. Our algorithm significantly improves the performance and scalability of decision-focused learning, and is scalable to real-world RMAB problem sizes.

\section*{Acknowledgments}
Aditya Mate was supported by the ARO and was accomplished under Grant Number W911NF-17-1-0370. Sanket Shah and Kai Wang were also supported by W911NF-17-1-0370 and ARO Grant Number W911NF-18-1-0208. The views and conclusions contained in this document are those of the authors and should not be interpreted as representing the official policies, either expressed or implied, of ARO or the U.S. Government. The U.S. Government is authorized to reproduce and distribute reprints for Government purposes notwithstanding any copyright notation herein.
Kai Wang~was additionally supported by Siebel Scholars.

\bibliography{reference}

\newpage
\appendix
\onecolumn
\section*{Appendix}

\section{Hyperparameter Setting and Computation Infrastructure}\label{sec:computation-infrastructure}
We run both Decision Focused Learning and Two-Stage Learning for 50 epochs in 2-state and 5-state synthetic domain problems, 30 epochs in ARMMAN domain and 18 epochs in 2-state partially observable setting. The learning rate $r$ is kept at $0.01$ and $\gamma=0.59$ is used in all experiments. 
All the experiments are performed on an Intel Xeon CPU with 64 cores and 128 GB memory.

\subsection*{Neural Network Structure}
The predictive model $\model_\weight$ we use to predict the transition probability is a neural network with an intermediate layer of size $64$ with ReLU activation function, and an output layer of size of the transition probability followed by a softmax layer to match probability distribution. Dropout layers are added to avoid overfitting. The same neural network structure is applied to all domains and all training methods.

In the synthetic datasets, given the generated transition probabilities, we feed the transition probability of each arm into a randomly initialized neural network with two intermediate layers each with $64$ neurons, and an output dimension size $16$ to generate a feature vector of size $16$.
The randomly initiated neural network uses ReLU layers as nonlinearity followed by a linear layer in the end.


\section{Real ARMMAN Dataset}\label{sec:dataset}
The large-scale quality improvement study conducted by~\citet{armman} contains 7668 beneficiries in the Round Robin Group. Over a duration of 7 weeks, $20\%$ of the beneficiaries receive at least one active action (LIVE service call). We randomly split the 7668 beneficiaries into 12 groups while preserving the proportion of beneficiaries who received at least one active action. There are 43 features available for every beneficiary which describe characteristics such as age, income, education level, call slot preference, language preference, phone ownership etc.

\subsection{Protected and Sensitive Features}
ARMMAN’s mobile voice call program has long been working with socially disadvantaged populations. ARMMAN does not collect or include constitutionally protected and particularly sensitive categories such as caste and religion. Despite such categories not being available, in pursuit of ensuring fairness, we worked with public health and field experts to ensure indicators such as education, and income levels that signify markers of socio-economic marginalization were measured and evaluated for fairness testing.

\subsection{Feature List}
We provide the full list of 43 features used for predicting transition probability:

\begin{itemize}
    \item Enroll gestation age, age (split into 5 categories), income (8 categories), education level (7 categories), language (5 categories), phone ownership (3 categories), call slot preference (5 categories), enrollment channel (3 categories), stage of pregnancy, days since first call, gravidity, parity, stillbirths, live births
\end{itemize}

\subsection{Feature Evaluation}

\begin{table}[ht]
\centering
\begin{tabular}{|l|l|l|l|}
\hline
\textbf{Feature} & \textbf{Two-stage} & \textbf{Decision-focused learning} & \textbf{p-value} \\ \hline
age (year)       & 25.57              & 24.9                               & 0.06             \\ \hline
gestation age (week) & 24.28              & 17.21                              & 0.00             \\ \hline
\end{tabular}
\caption{Feature analysis of continuous features. This table summarizes the average feature values of the beneficiaries selected to schedule service calls by different learning methods. The p-value of the continuous features is analyzed using t-test for difference in mean.}
\label{table:p-value-continuous-feature}
\end{table}

\begin{table}[ht]
\centering
\begin{tabular}{|l|lll|}
\hline
\textbf{Feature}          & \multicolumn{1}{l|}{\textbf{Two-stage}} & \multicolumn{1}{l|}{\textbf{Decision-focused learning}} & \textbf{p-value} \\ \hline
income (rupee, averaged over multiple categories)            & \multicolumn{1}{l|}{10560.0}            & \multicolumn{1}{l|}{11190.0}           & 0.20             \\ \hline
education (categorical)   & \multicolumn{1}{l|}{3.32}   & \multicolumn{1}{l|}{3.16}              & 0.21             \\ \hline
stage of pregnancy        & \multicolumn{1}{l|}{0.13}               & \multicolumn{1}{l|}{0.03}                               & \textbf{0.00}    \\ \hline
language                  & \multicolumn{3}{l|}{}                                                                                                \\ \hline
language (hindi)          & \multicolumn{1}{l|}{0.53}               & \multicolumn{1}{l|}{0.6}                                & \textbf{0.04}    \\ \hline
language (marathi)        & \multicolumn{1}{l|}{0.45}               & \multicolumn{1}{l|}{0.4}                                & 0.08             \\ \hline
phone ownership           & \multicolumn{3}{l|}{}                                                                                                \\ \hline
phone ownership (women)   & \multicolumn{1}{l|}{0.86}               & \multicolumn{1}{l|}{0.82}                               & \textbf{0.04}    \\ \hline
phone ownership (husband) & \multicolumn{1}{l|}{0.12}               & \multicolumn{1}{l|}{0.16}                               & \textbf{0.03}    \\ \hline
phone ownership (family)  & \multicolumn{1}{l|}{0.02}               & \multicolumn{1}{l|}{0.02}                               & 1.00             \\ \hline
enrollment channel        & \multicolumn{3}{l|}{}                                                                                                \\ \hline
channel type (community)  & \multicolumn{1}{l|}{0.7}                & \multicolumn{1}{l|}{0.47}                               & \textbf{0.00}    \\ \hline
channel type (hospital)   & \multicolumn{1}{l|}{0.3}                & \multicolumn{1}{l|}{0.53}                               & \textbf{0.00}    \\ \hline
\end{tabular}
\caption{Feature analysis of categorical features. This table summarizes the average feature values of the beneficiaries selected to schedule service calls by different learning methods. The p-value of the categorical values is analyzed using chi-square test for different proportions.}
\label{table:p-value-categorical-feature}
\end{table}
In our simulation, we further analyze the demographic features of participants who are selected to schedule service calls by either two-stage learning method and decision-focused learning method. The following tables show the average value of each individual feature over the selected participants with scheduled service calls under the two-stage or decision-focused learning method. The p-value of the continuous features is analyzed using t-test for difference in mean; the p-value of the categorical values is analyzed using chi-square test for different proportions. 

In Table~\ref{table:p-value-continuous-feature} and Table~\ref{table:p-value-categorical-feature}, we can see that there is no statistical significance (p-value $> 0.05$) between the average feature values of income and education, meaning that there is no obvious difference in these feature values between the population selected by two different methods. We see statistical significance in some other features, e.g., gestation age, stage of maternal event, language, phone ownership, and channel type, which may be further analyzed to understand the benefit of decision-focused learning, but they do not appear to directly bear upon socio-economic marginalization; these features are more related to the health status of the beneficiaries.

\section{Consent for Data Collection and Analysis}\label{sec:consent}
In this section, we provide information about consent related to data collection, analyzing data, data usage and sharing.


\subsection{Secondary Analysis and Data Usage}
This study falls into the category of secondary analysis of the aforementioned dataset. We use the previously collected engagement trajectories of different beneficiaries participating in the service call program to train the predictive model and evaluate the performance.
The evaluation of the proposed algorithm is evaluated via different off-policy policy evaluations, including an importance sampling-based method and a simulation-based method discussed in Section~\ref{sec:policy-evaluation}.
This paper does not involve deployment of the proposed algorithm or any other baselines to the service call program.As noted earlier, the experiments are secondary analysis using different evaluation metrics with approval from the ARMMAN ethics board.

\subsection{Consent for Data Collection and Sharing}
The consent for collecting data is obtained from each of the participants of the service call program. The data collection process is carefully explained to the participants to seek their consent before collecting the data. The data is anonymized before sharing with us to ensure anonymity. 
Data exchange and use was regulated through clearly defined exchange protocols including anonymization, read-access only to researchers, restricted use of the data for research purposes only, and  approval by ARMMAN's ethics review committee.

\subsection{Universal Accessibility of Health Information}
To allay further concerns: this simulation study focuses on improving quality of service calls. Even in the intended future application, all participants will receive the same weekly health information by automated message regardless of whether they are scheduled to receive service calls or not.  The service call program does not withhold any information from the participants nor conduct any experimentation on the health information. The health information is always available to all participants, and participants can always request service calls via a free missed call service. In the intended future application our algorithm may only help schedule *additional*  service calls to help beneficiaries who are likely to drop out of the program.

\section{Societal Impacts and Limitations}
\subsection{Societal Impacts}
The improvement shown in the real dataset directly reflects the number of engagements improved by our algorithm under different evaluation metrics. On the other hand, because of the use of demographic features to predict the engagement behavior, we must carefully compare the models learned by standard two-stage approach and our decision-focused learning to further examine whether there is any bias or discrimination concern.

Specifically, the data is collected by ARMMAN, an India non-government organization, to help mothers during their pregnancy.
The ARMMAN dataset we use in the paper does not contain information related to race, religion, caste or other sensitive features; this information is not available to the machine learning algorithm. Furthermore, examination by ARMMAN staff of the mothers selected for service calls by our algorithm did not reveal any specific bias related to these features. In particular, the program run by ARMMAN targets mothers in economically disadvantaged communities; the majority of the participants (94\%) are below the international poverty line determined by The World Bank~\cite{world2020poverty}.
To compare the models learned by two-stage and DF-Whittle approach, we further examine the difference between those mothers who are selected for service call in two-stage and DF-Whittle, respectively. We observe that there are some interesting differences. For example, DF-Whittle chooses to do service calls to expectant mothers earlier in gestational age (22\% vs 37\%), and to a lower proportion of those who have already given birth (2.8\% vs 13\%) compared to two-stage, but in terms of the income level, 94\% of the mothers selected by both methods are below the poverty line. This suggests that our approach is not biased based on income level, especially when the entire population is coming from economically disadvantaged communities. Our model can identify other features of mothers who are actually in need of service calls.

\subsection{Limitations}

\paragraph{Impact of limited data and the strength of decision-focused learning}
As shown in Section~\ref{sec:results} and Figure~\ref{fig:trajectory}, we notice a smaller improvement between decision-focused learning and two-stage approach when there is sufficient data available in the training set. This is because the data is sufficient enough to train a predictive model with small predictive loss, which implies that the predicted transition probabilities and the true transition probabilities are also close enough with similar Whittle indices and Whittle index policy. In this case with sufficient data, there is less discrepancy between predictive loss and the evaluation metrics, which suggests less improvement led by fixing the discrepancy using decision-focused learning.
Compared to two-stage approach, decision-focused learning is still more expensive to run.
Therefore, when data is sufficient, two-stage may be sufficient to achieve comparable performance while maintaining a low training cost.

On the other hand, we notice a larger improvement between decision-focused learning and two-stage approach when data is limited. When data is limited, predictive loss is less representative with a larger mismatch compared to the evaluation metrics. Therefore, fixing the objective mismatch issue using decision-focused learning becomes more prominent. Therefore, decision-focused learning may be adopted in the limited data case to significantly improve the performance.

\paragraph{Computation cost}
As we have shown in Section~\ref{sec:computation-cost}, our approach improves the computation cost of decision-focused learning from $O(M^{\omega N})$ to $O(NM^{\omega + 1})$, where $N$ is the number of arms and $M$ is the number of states.
This computation cost is linear in the number of arms $N$, allowing us to scale up to large real-world deployment of RMAB applications with larger number of arms involved in the problem.
Nonetheless, the extension in terms of the number of states $M$ is not cheap. The computation cost still grows between cubic and biquadratic as shown in Figure~\ref{fig:computation-cost-states}.
This is particularly significant when working on partially observable RMAB problems, where the partially observable problems are reduced to fully observable problems with larger number of states.
There is room for improving the computation cost in terms of the number of states to make decision-focused learning more scalable to real-world applications.


\section{Computation Cost Analysis of Decision-focused Learning}\label{sec:computation-cost-appendix}

We have shown the computation cost of backpropagating through Whittle indices in Section~\ref{sec:computation-cost}. This section covers the remaining computation cost associated to other components, including the computation cost of Whittle indices in the forward pass, and the computation cost of constructing soft Whittle index policy using soft-top-k operator.

\subsection{Solving Whittle Index (Forward Pass)}\label{sec:computation-cost-whittle-forward}
In this section, we discuss the cost of computing Whittle index in the forward pass. In the work by~\citet{qian2016restless}, they propose to use value iteration and binary search to solve the Bellman equation with $\nstates$ states. Therefore, every value iteration requires updating the current value functions of $\nstates$ states by considering all the possible $\nstates^2$ transitions between states, which results in a computation cost of $O(\nstates^2)$ per value iteration.
The value iteration is run for a constant number of iterations, and the binary search is run for $O(\log \frac{1}{\epsilon})$ iterations to get a precision of order $\epsilon$. In total, the computation cost is of order $O(\nstates^2 \log \frac{1}{\epsilon}) = O(\nstates^2)$ where we simply use a fixed precision to ignore the dependency on $\epsilon$.

On the other hand, there is a faster way to compute the value function by solving linear program with $\nstates$ variables directly. The Bellman equation can be expressed as a linear program where all the $\nstates$ variables are the value functions. The best known complexity of solving a linear program with $\nstates$ variables is $O(M^{2 + \frac{1}{18}})$ by \citet{jiang2020faster}. Notice that this complexity is slightly larger than the one in value iteration because (i) value iteration does not guarantee convergence in a constant iterations (ii) the constant associated to the number of value iterations is large.

In total, we need to compute the Whittle index of $N$ arms and for $\nstates$ possible states in $\State$. The total complexity of value iteration and linear program are $O(N \nstates^3)$ with a large constant and $O(N \nstates^{3 + \frac{1}{18}})$, respectively.
In any cases, the cost of computing all Whittle indices in the forward pass is still smaller than $O(N \nstates^{1 + \omega})$, the cost of backpropagating through all the Whittle indices in the backward pass.
Therefore, the backward pass is the bottleneck of the entire process.

\subsection{Soft-top-k Operators}
In Section~\ref{sec:computation-cost-whittle-forward} and Section~\ref{sec:computation-cost}, we analyze the cost of computing and backpropagating through Whittle indices of all states and all arms. In this section, we discuss the cost of computing the soft Whittle index policy from the given Whittle indices using soft-top-k operators.

\paragraph{Soft-top-k operators}
\citet{xie2020differentiable} reduces top-k selection problem to an optimal transport problem that transports a uniform distribution across all input elements with size $N$ to a distribution where the elements with the highest-k values are assigned probability $1$ and all the others are assigned $0$.

This optimal transport problem with $N$ elements can be efficiently solved by using Bregman projections~\cite{benamou2015iterative} with complexity $O(L N)$, where $L$ is the number of iterations used to run Bregman projections.
In the backward pass, \citet{xie2020differentiable} shows that the technique of differentiating through the fixed point equation~\cite{bai2019deep,amos2017optnet} also applies, but the naive implementation requires computation cost $O(N^2)$. Therefore, ~\citet{xie2020differentiable} provides a faster computation approach by leveraging the associate rule in matrix multiplication to lower the backward complexity to $O(N)$.

In summary, a single soft-top-k operator requires $O(LN)$ to compute the result in the forward pass, and $O(N)$ to compute the derivative in the backward pass.
In our case, we need to apply one soft-top-k operator for every time step in $T$ and for every trajectory in $\trajectories$. Therefore, the total computation cost of computing a soft Whittle index policy and the associated importance sampling-based evaluation metric is bounded by $O(L N T |\trajectories|)$, which is linear in the number of arms $N$, but still significantly smaller than $O(N \nstates^{\omega + 1})$, the cost of backpropagating through all Whittle indices as shown in Section~\ref{sec:computation-cost}.
Therefore, we just need to concern the computation cost of Whittle indices in decision-focused learning.

\subsection{Computation Cost Dependency on the Number of States}
\begin{figure}
    \centering
    \includegraphics[width=0.4\textwidth]{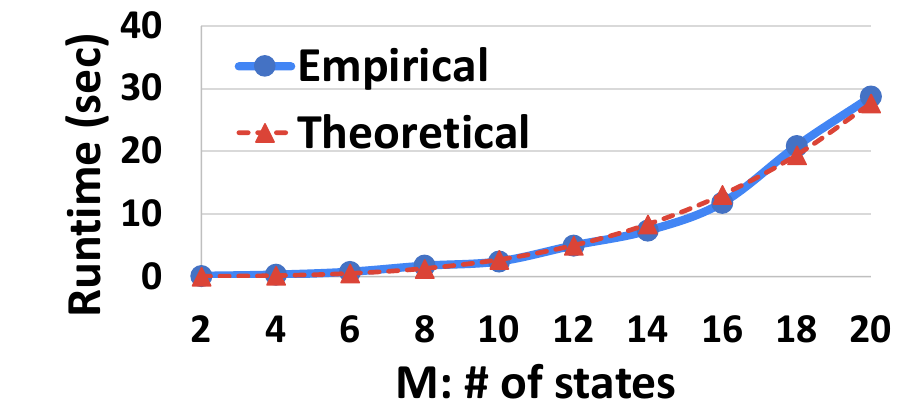}
    \caption{Computation cost comparison to the theoretical guarantee with varying number of states $M$.}
    \label{fig:computation-cost-states}
\end{figure}

Figure~\ref{fig:computation-cost-states} compares the computation cost of our algorithm, DF-Whittle, and the theoretical computation cost $O(N M^{\omega + 1}$. We vary the number of states $M$ in Figure~\ref{fig:computation-cost-states} and we can see that the computation cost of our algorithm matches the theoretical guarantee on the computation cost. In contrast to the prior work with computation cost $O(M^{\omega N})$, our algorithm significantly improves the computation cost of running decision-focused learning on RMAB problems.

\section{Importance Sampling-based Evaluations for ARMMAN Dataset with Single Trajectory}\label{sec:is-single-trajectory}
Unlike the synthetic datasets that we can produce multiple trajectories of an RMAB problem, in the real problem of service call scheduling problem operated by ARMMAN, there is only one trajectory available to us for every RMAB problem. Due to the specialty of the maternal and child health domain, it is unlikely to have the exactly same set of the pregnant mothers participating in the service call scheduling program at different times and under the same engagement behavior.

Given this restriction, we must evaluate the performance of a newly proposed policy using the only available trajectory. Unfortunately, the standard CWPDIS in Equation~\ref{eqn:IS-OPE} does not work because the CWPDIS estimator is canceled out when there is only one trajectory:

\begin{align}
    \text{Eval}_\text{IS}(\policy, \trajectories) = \sum\limits_{t \in [T], i \in [N]} \gamma^{t-1} \frac{\mathop{\mathbb{E}}_{\trajectory \sim \trajectories} \left[ \reward_{t,i} \rho_{ti}(\trajectory) \right]}{\mathop{\mathbb{E}}_{\trajectory \sim \trajectories} \left[ \rho_{ti}(\trajectory) \right]} = \sum\limits_{t \in [T], i \in [N]} \gamma^{t-1} \frac{\reward_{t,i} \rho_{ti}(\trajectory)}{\rho_{ti}(\trajectory)} = \sum\limits_{t \in [T], i \in [N]} \gamma^{t-1} \reward_{t,i}
\end{align}
which is fixed regardless what target policy $\pi$ is used and the associated importance sampling weights $\frac{\policy(\action_{t,i} \mid \state_t)}{\policy_{\text{beh}}(\action_{t,i} \mid \state_t)}$ and $\rho_{ti} = \prod\nolimits_{t'=1}^t \frac{\policy(\action_{t',i} \mid \state_{t'})}{\policy_{\text{beh}}(\action_{t',i} \mid \state_{t'})}$.
This implies that we cannot use CWPDIS to evaluate the target policy when there is only one trajectory. 

Accordingly, we use the following variant to evaluate the performance:
\begin{align}\label{eqn:eqn:IS-OPE-single-trajectory}
    \text{Eval}_\text{IS}(\policy, \trajectories) = \sum\limits_{i \in [N], t \in [T]} \gamma^{t-1} \frac{\reward_{t,i} \rho'_{ti}(\trajectory)}{\mathop{\mathbb{E}}\nolimits_{t' \in [T]} \left[ \rho'_{t' i}(\trajectory) \right]}
\end{align}
where the new importance sampling weights are defined by $\rho'_{t,i}(\trajectory) = \frac{\policy(\action_{t,i} \mid \state_t)}{\policy_{\text{beh}}(\action_{t,i} \mid \state_t)}$, which is not multiplicative compared to the original ones.

The main motivation of this new evaluation metric is to segment the given trajectory into a set of length-1 trajectories.
We can apply CWPDIS to the newly generated length-1 trajectories to compute a meaningful estimate because we have more than one trajectory now.
The OPE formulation with segmentation is under the assumption that we can decompose the total reward into the contribution of multiple segments using the idea of trajectory segmentation~\cite{krishnan2017transition,ranchod2015nonparametric}. This assumption holds when all segments start with the same state distribution. In our ARMMAN dataset, the data is composed of trajectories of the participants who have enrolled in the system a few weeks ago, which have (almost) reached a stationary distribution. Therefore, the state distribution under the behavior policy, which is a uniform random policy, does not change over time. Our assumption of identical distribution is satisfied and we can decompose the trajectories into smaller segments to perform evaluation.
Empirically, we noticed that this temporal decomposition helps define a meaningful importance sampling-based evaluation with the consistency benefit brought by CWPDIS.

\section{Additional Experimental Results}\label{sec:additional-results}
\begin{figure}[ht]
    \centering
    \begin{subfigure}{0.325\linewidth}
        \centering
        \includegraphics[width=\textwidth]{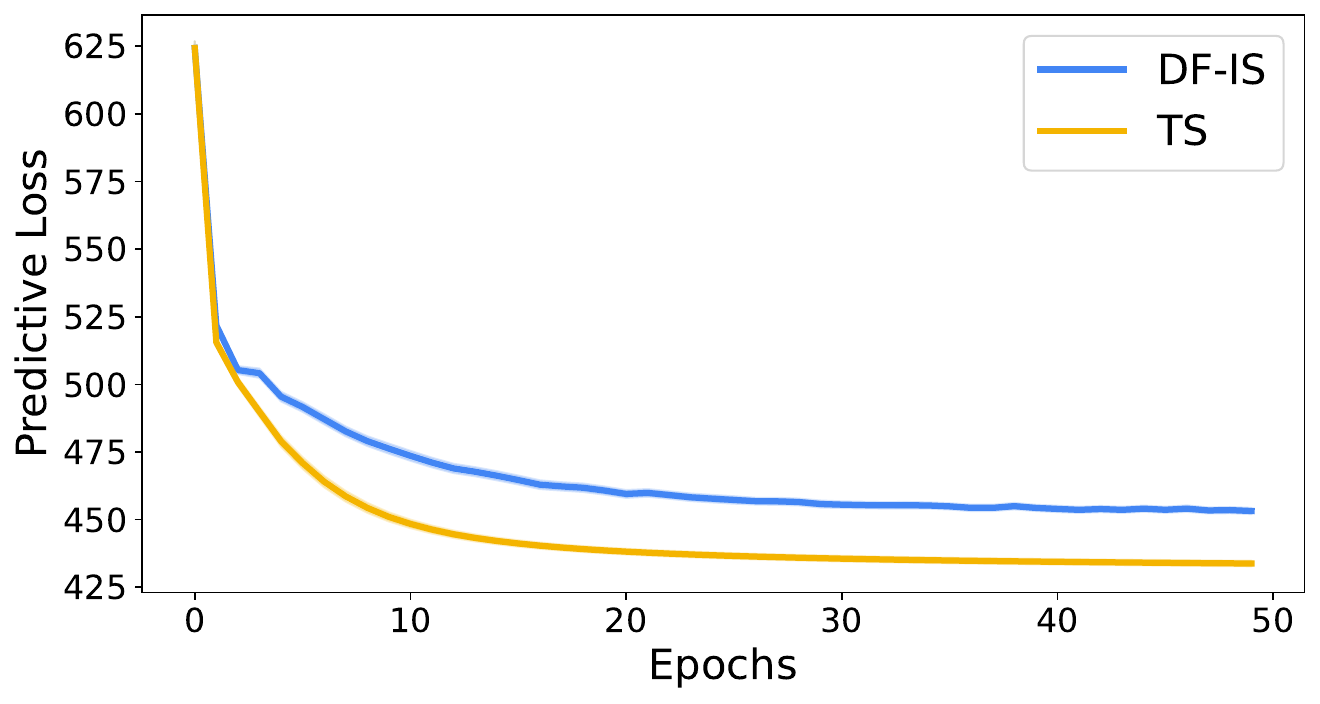}
        \caption{Testing predictive loss v.s. training epoch}
        \label{fig:learning-curve-synth-loss}
    \end{subfigure}
    \hfill
    \begin{subfigure}{0.325\linewidth}
        \centering
        \includegraphics[width=\textwidth]{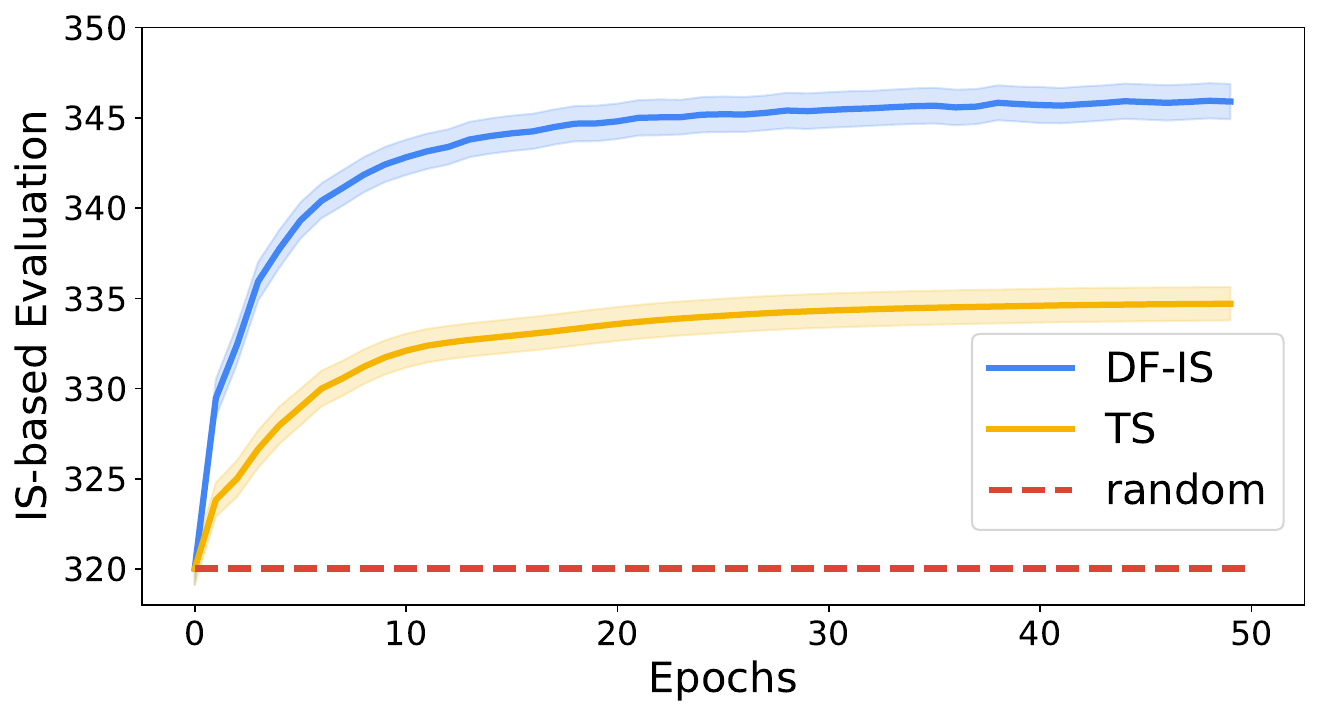}
        \caption{Testing IS-based evaluation}
        \label{fig:learning-curve-synth-is}
    \end{subfigure}
    \hfill
    \begin{subfigure}{0.325\linewidth}
        \centering
        \includegraphics[width=\textwidth]{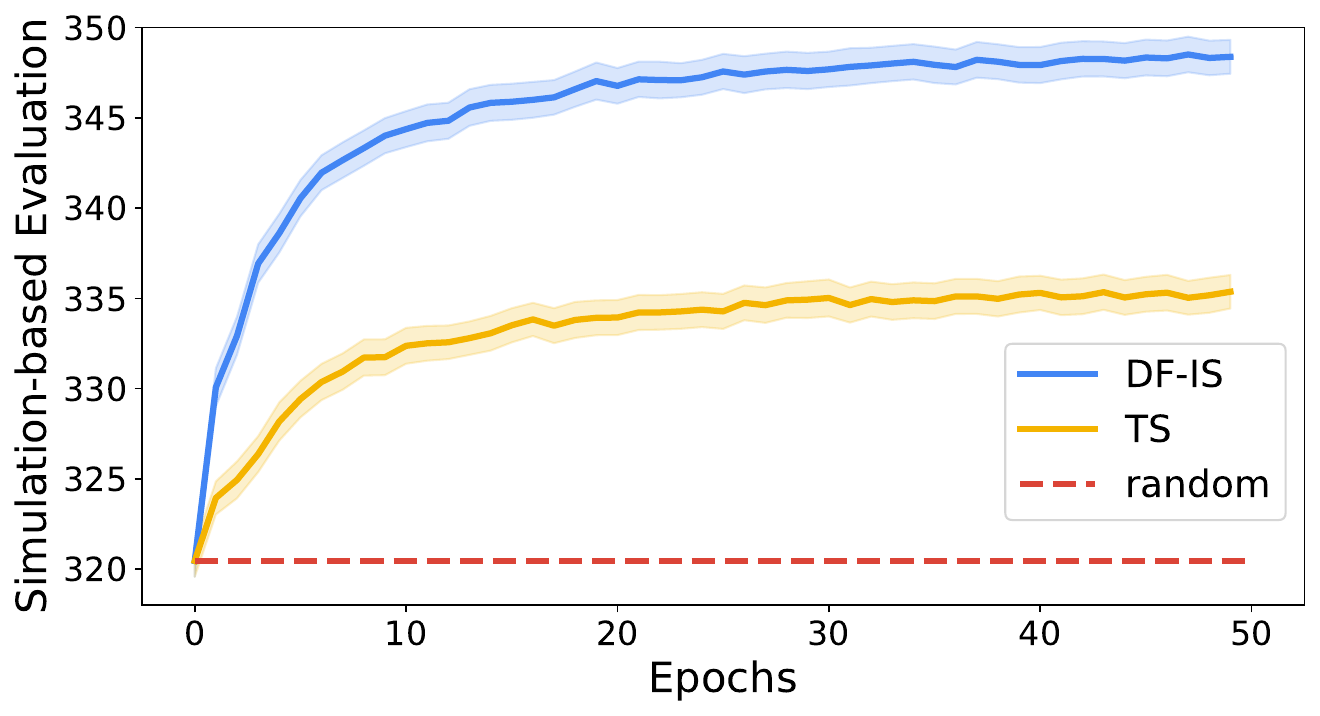}
        \caption{Testing simulation-based evaluation}
        \label{fig:learning-curve-synth-sim}
    \end{subfigure}
    \caption{Comparison between two-stage and decision-focused in the synthetic fully observable 2-state RMAB problems. 
    }
    \label{fig:learning-curve-synth}
\end{figure}

\begin{figure}[ht]
    \centering
    \begin{subfigure}{0.325\linewidth}
        \centering
        \includegraphics[width=\textwidth]{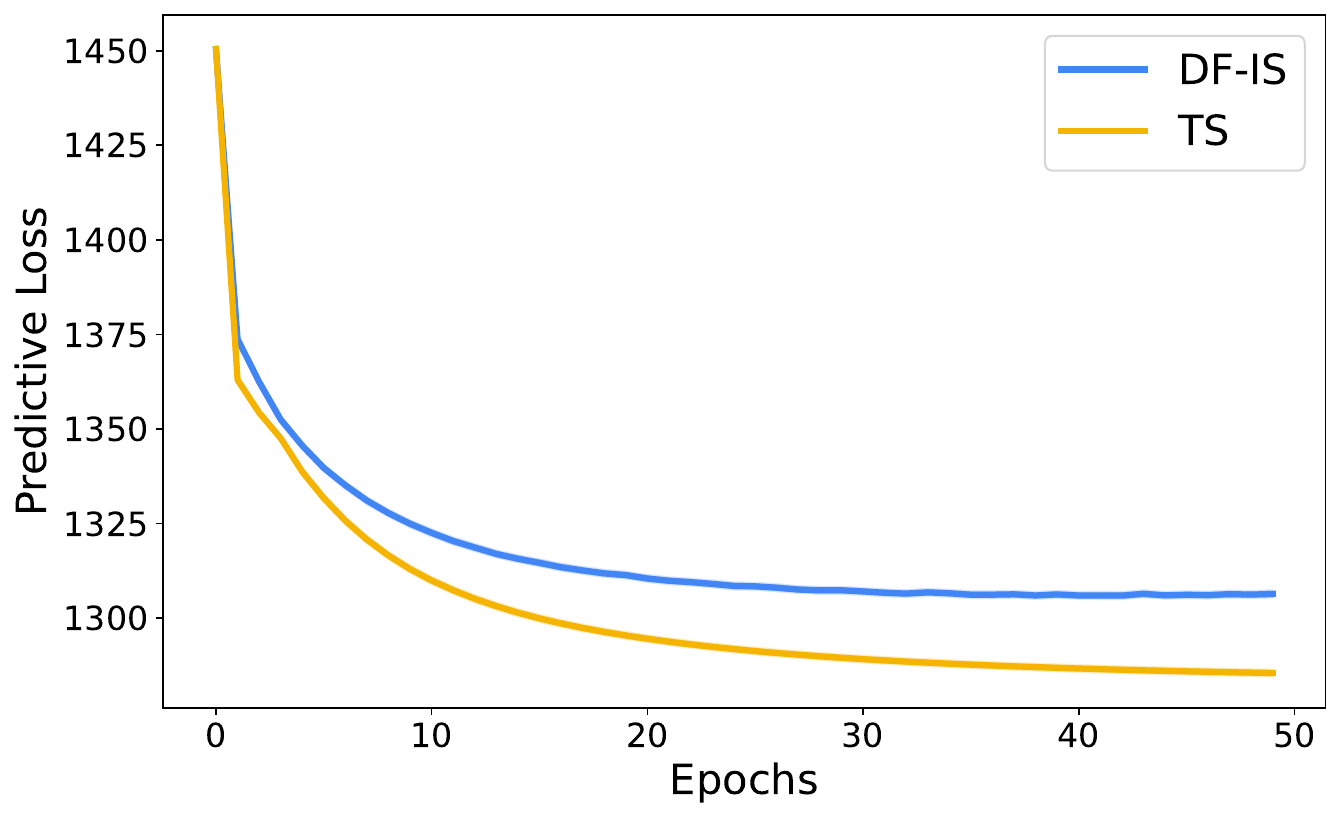}
        \caption{Testing predictive loss v.s. training epoch}
        \label{fig:learning-curve-5-state-loss}
    \end{subfigure}
    \hfill
    \begin{subfigure}{0.325\linewidth}
        \centering
        \includegraphics[width=\textwidth]{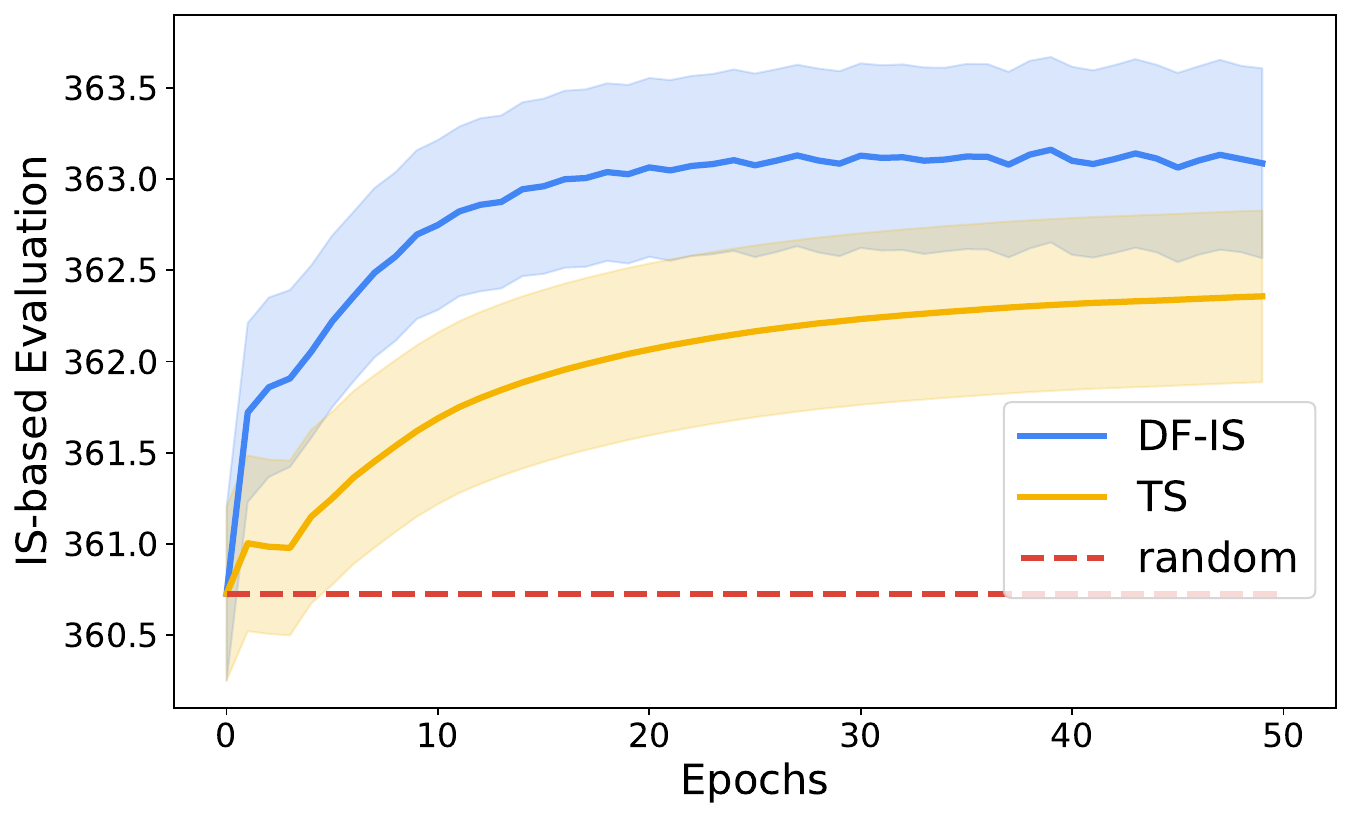}
        \caption{Testing IS-based evaluation}
        \label{fig:learning-curve-5-state-is}
    \end{subfigure}
    \hfill
    \begin{subfigure}{0.325\linewidth}
        \centering
        \includegraphics[width=\textwidth]{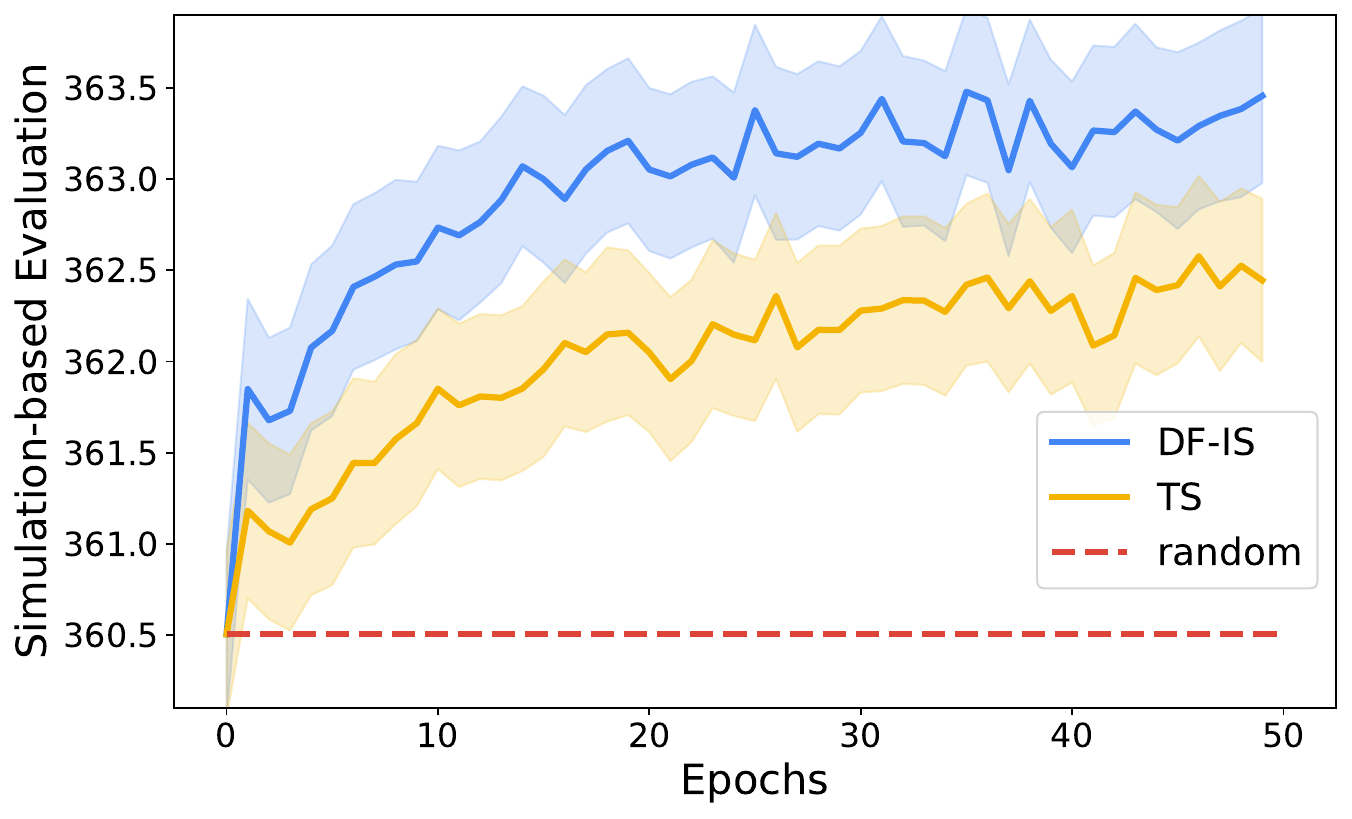}
        \caption{Testing simulation-based evaluation}
        \label{fig:learning-curve-5-state-sim}
    \end{subfigure}
    \caption{Comparison between two-stage and decision-focused learning for fully observable 5-state RMAB problems.}
    \label{fig:learning-curve-5-state}
\end{figure}

\begin{figure}[ht]
    \centering
    \begin{subfigure}{0.325\linewidth}
        \centering
        \includegraphics[width=\textwidth]{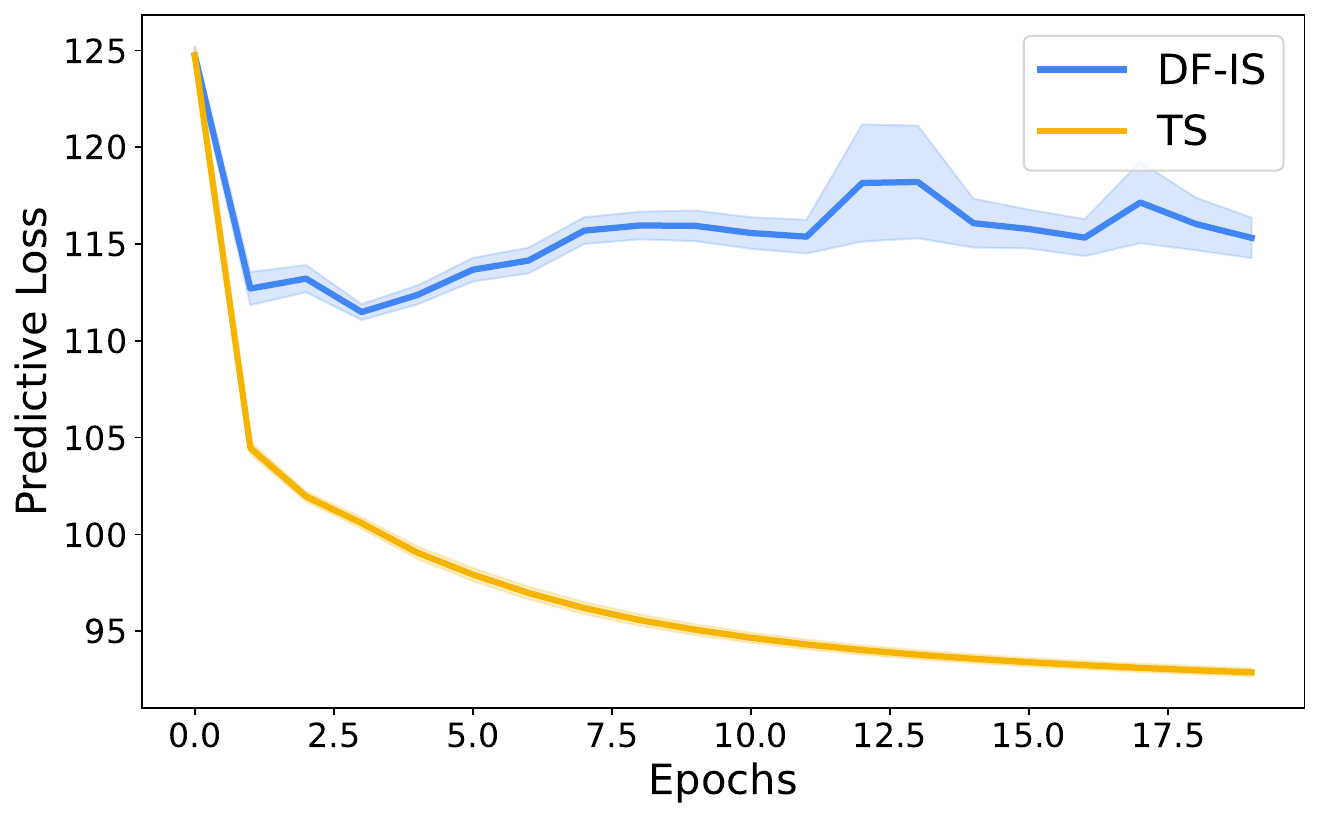}
        \caption{Testing predictive loss v.s. training epoch}
        \label{fig:learning-curve-2-state-partial-loss}
    \end{subfigure}
    \hfill
    \begin{subfigure}{0.325\linewidth}
        \centering
        \includegraphics[width=\textwidth]{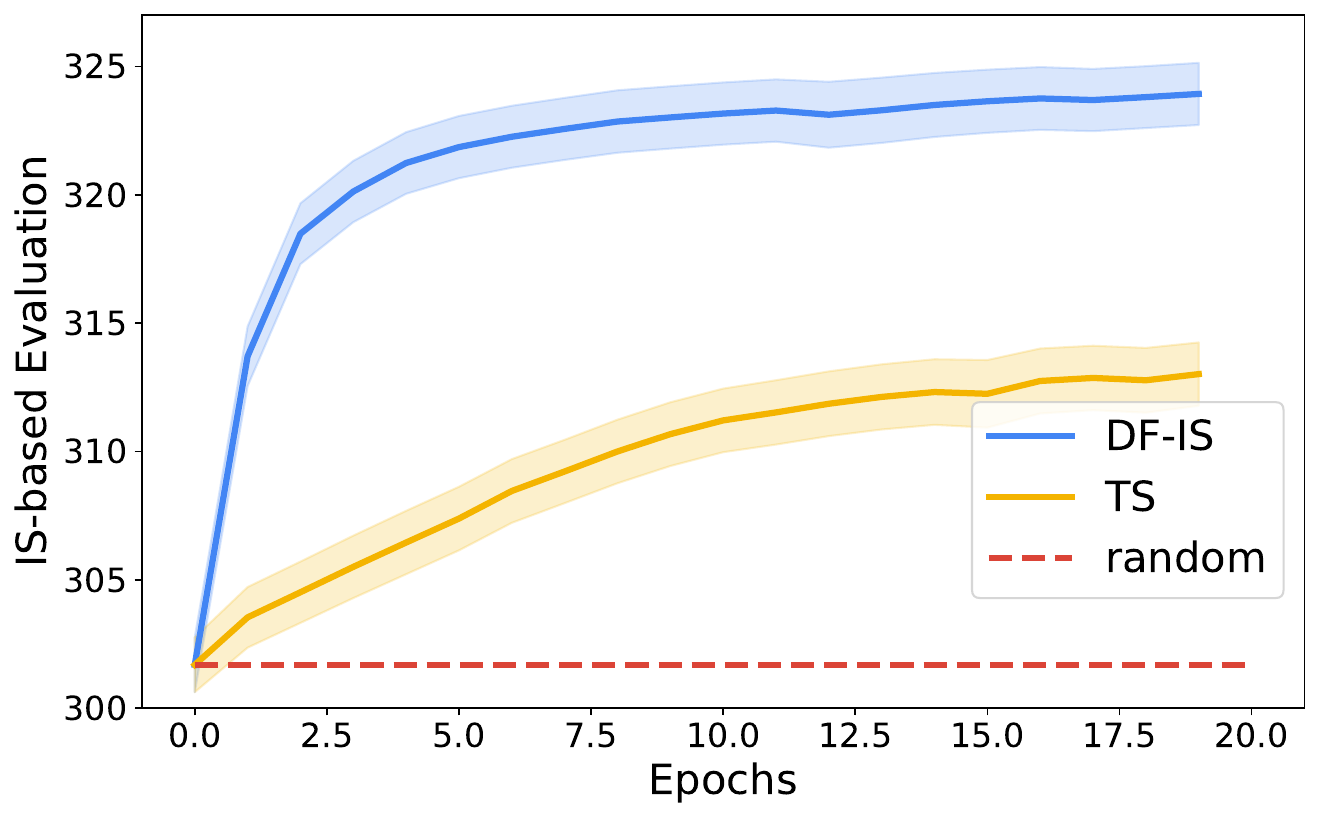}
        \caption{Testing IS-based evaluation}
        \label{fig:learning-curve-2-state-partial-is}
    \end{subfigure}
    \hfill
    \begin{subfigure}{0.325\linewidth}
        \centering
        \includegraphics[width=\textwidth]{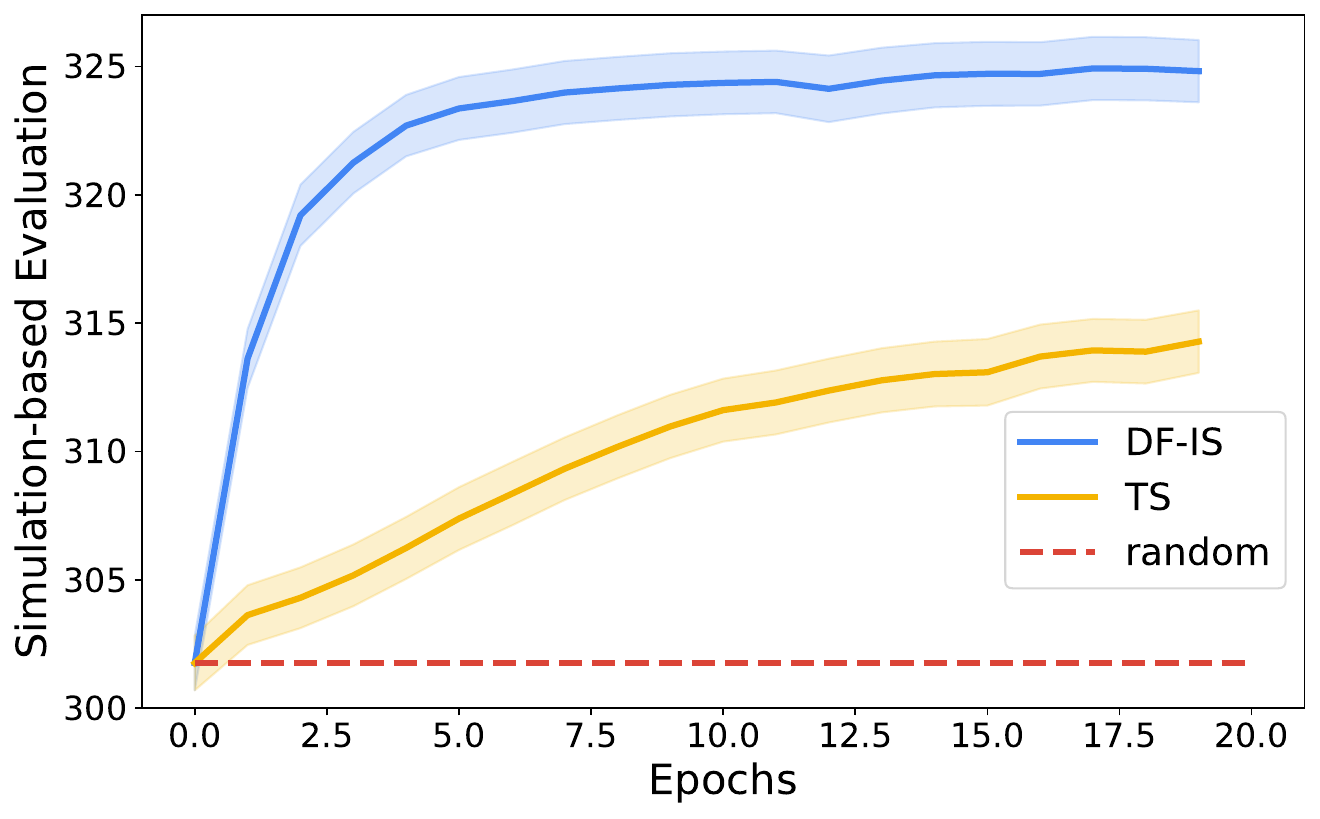}
        \caption{Testing simulation-based evaluation}
        \label{fig:learning-curve-2-state-partial-sim}
    \end{subfigure}
    \caption{Comparison between two-stage and decision-focused learning for 2-state partially observable RMAB problems.}
    \label{fig:learning-curve-2-state-partial}
\end{figure}

\begin{figure}[ht]
    \centering
    \begin{subfigure}{0.325\linewidth}
        \centering
        \includegraphics[width=\textwidth]{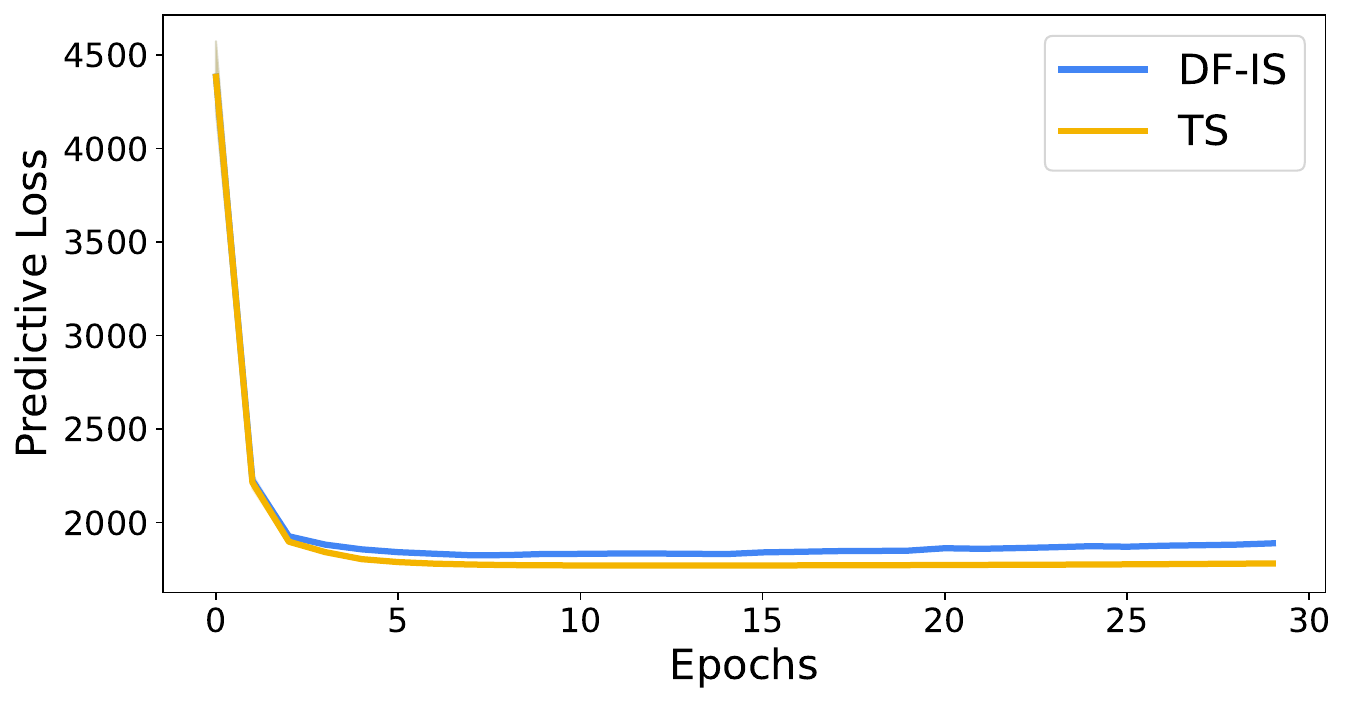}
        \caption{Testing predictive loss}
        \label{fig:learning-curve-armman-loss}
    \end{subfigure}
    \hfill
    \begin{subfigure}{0.325\linewidth}
        \centering
        \includegraphics[width=\textwidth]{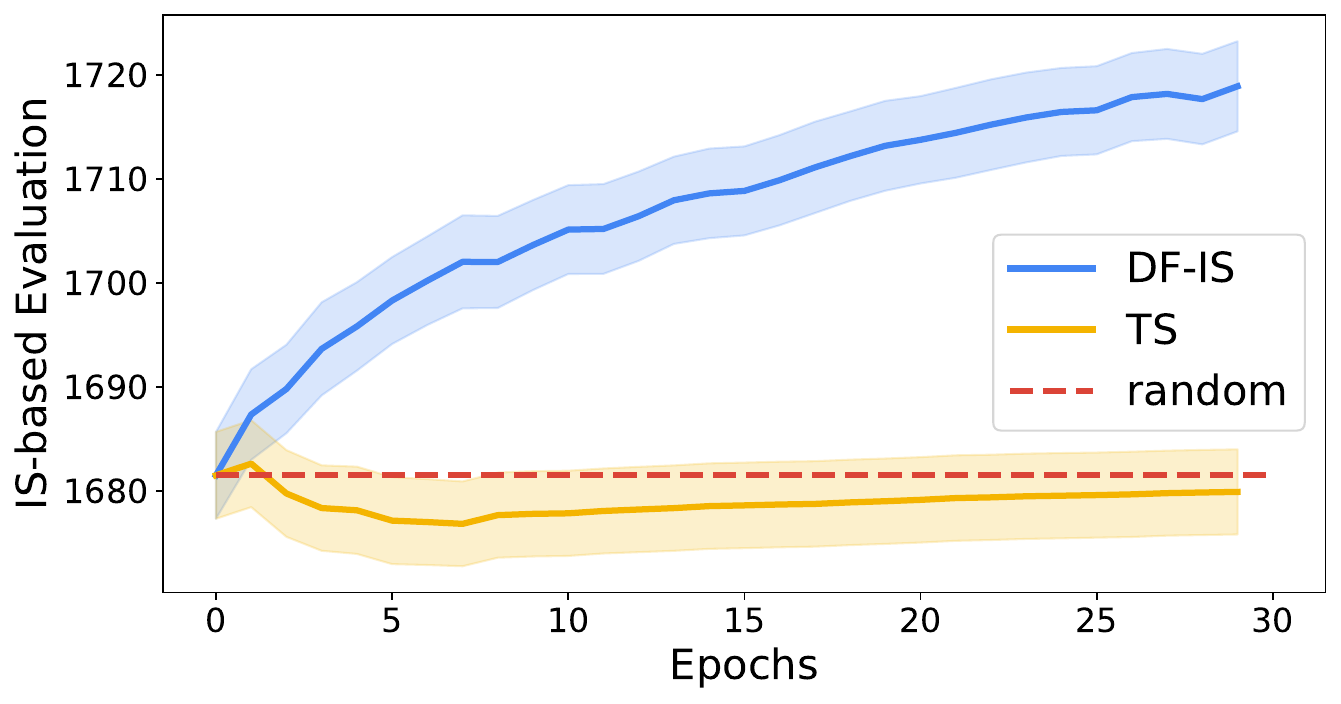}
        \caption{Testing IS-based evaluation}
        \label{fig:learning-curve-armman-is}
    \end{subfigure}
    \hfill
    \begin{subfigure}{0.325\linewidth}
        \centering
        \includegraphics[width=\textwidth]{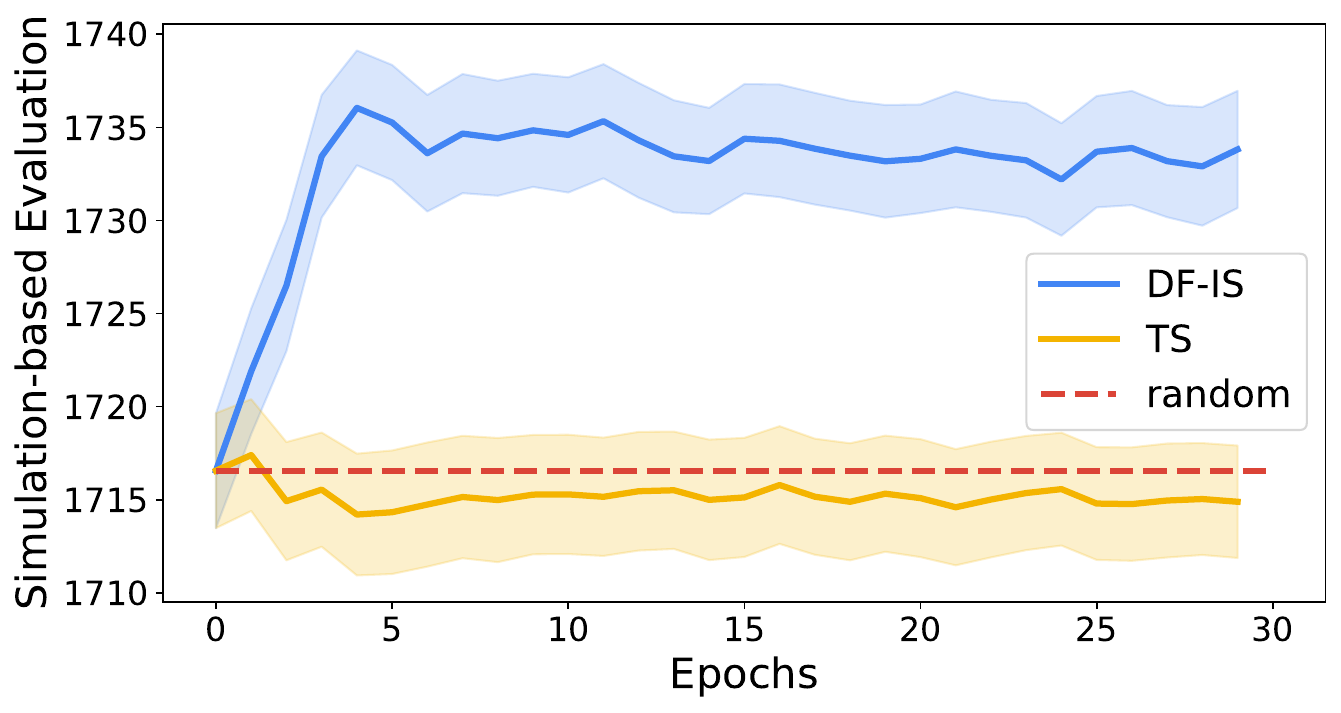}
        \caption{Testing simulation-based evaluation}
        \label{fig:learning-curve-armman-sim}
    \end{subfigure}
    \caption{Comparison between two-stage and decision-focused learning in the real ARMMAN service call scheduling problem. The pulling action in the real dataset is much sparser, leading to a larger mismatch between predictive loss and evaluation metrics. Two-stage overfits to the predictive loss drastically with no improvement in evaluation metrics. In contrast, decision-focused learning can directly optimize the evaluation metric to avoid the objective mismatch issue.
    }
    \label{fig:learning-curve-armman}
\end{figure}

We provide the learning curves of fully observable 2-state RMAB, fully observable 5-state RMAB, partially observable 2-state RMAB, and the real ARMMAN fully observable 2-state RMAB problems in Figure~\ref{fig:learning-curve-synth},\ref{fig:learning-curve-5-state},~\ref{fig:learning-curve-2-state-partial},~\ref{fig:learning-curve-armman}, respectively.
Across all domains, two-stage method consistently converges to a lower predictive loss faster than decision-focused learning in Figure~\ref{fig:learning-curve-synth-loss},\ref{fig:learning-curve-5-state-loss},~\ref{fig:learning-curve-2-state-partial-loss},~\ref{fig:learning-curve-armman-loss}.
However, the learned model does not produce a policy with good performance in the importance sampling-based evaluation metric in Figure~\ref{fig:learning-curve-synth-is},\ref{fig:learning-curve-5-state-is},~\ref{fig:learning-curve-2-state-partial-is},~\ref{fig:learning-curve-armman-is}, and similarly in the simulation-based evaluation metric in Figure~\ref{fig:learning-curve-synth-sim},\ref{fig:learning-curve-5-state-sim},~\ref{fig:learning-curve-2-state-partial-sim},~\ref{fig:learning-curve-armman-sim}.

\section{Solving for and Differentiating Through the Whittle Index Computation}\label{sec:differentiability-details}

\noindent To solve for the Whittle index for some state $\tmpstate \in S$, you have to solve the following set of equations:
\begin{align}
    V(\tmpstate) &= R(s) + m_{\tmpstate} + \gamma \sum_{\mathclap{s' \in \mathcal{S}}} P(s, 0, s') \cdot V(s') \nonumber \\
    V(\tmpstate) &= R(s) + \gamma \sum_{\mathclap{s' \in \mathcal{S}}} P(s, 1, s') \cdot V(s') \label{eqn:whittle} \\
    V(s) &= \max_{a \in \{0,1\}} [R(s) + (1-a)m_{\tmpstate} + \gamma \sum_{\mathclap{s' \in \mathcal{S}}} P(s, a, s') \cdot V(s')],\quad \quad \forall s \in \mathcal{S} - \tmpstate \label{eqn:bellman}
\end{align}
Here:
\begin{itemize}[align=parleft,labelsep=4em,leftmargin=6em]
    \item[$\mathcal{S}$] is the set of all states
    \item[$R(s)$] is the reward for being in state $s$
    \item[$P(s,a,s')$] is the probability of transitioning to state $s'$ when you begin in state $s$ and take action $a$
    \item[$V(s)$] is the expected value of being in state $s$
    \item[$m_s$] is the whittle index for state $s$
\end{itemize}

\vspace{1em}
\noindent One way to interpret these equations is to view them as the Bellman Optimality Equations associated with a modified MDP in which the reward function is changed to $R'_{\tmpstate}(s,a) = R(s) + (1-a)m_{\tmpstate}$, i.e., you are given a `subsidy' for not acting (Equation \ref{eqn:bellman}). Then, to find the whittle index for state $\tmpstate$, you have to find the minimum subsidy for which the value of not acting exceeds the value of acting \cite{whittle1988restless}. At this transition point, the value of not acting is equal to the value of acting in that state (Equation \ref{eqn:whittle}), leading to the set of equations above.

Now, this set of equations is typically hard to solve because of the $\max$ terms in Equation \ref{eqn:bellman}. Specifically, knowing whether $\arg\max_a = 0$ or $\arg\max_a = 1$ for some state $s$ is equivalent to knowing what the optimal policy is for this modified MDP; such equations are typically solved using Value Iteration or variations thereof. However, this problem is slightly more complicated than a standard MDP because one also has to determine the value of $m_{s}$. The way that this problem is traditionally solved in the literature is the following:
\begin{enumerate}[leftmargin=2.5em,topsep=0em,parsep=0em,itemsep=0em]
    \item One guesses a value for the subsidy $m_s$.
    \item Given this guess, one solves the Bellman Optimality Equations associated with the modified MDP.
    \item Then, one checks the resultant policy. If it is more valuable to act than to not act in state $s$, the value of the guess for the subsidy is increased. Else, it is decreased.
    \item Go to Step 2 and repeat until convergence.
\end{enumerate}
Given the monotonicity and the ability to bound the values of the whittle index, Step 3 above is typically solved using binary search. However, even with Binary Search, this process is quite time-consuming.

In this paper, we provide a much faster solution method for our application of interest. We leverage the small size of our state space to search over the space of policies rather than over the correct value of $m_s$. Concretely, because $S = \{0, 1\}$ and $A = \{0, 1\}$, the whittle index equations for state $s = 0$ above boil down to:
\begin{align}
    V(0) &= R(0) + m_{s_0} + \gamma \sum_{\mathclap{s' \in \{0, 1\}}} P(0, 0, s') \cdot V(s') \nonumber \\
    V(0) &= R(0) + \gamma \sum_{\mathclap{s' \in \{0, 1\}}} P(0, 1, s') \cdot V(s') \nonumber \\
    V(1) &= \max_{a \in \{0,1\}} [R(1) + (1-a)m_{s_0} + \gamma \sum_{\mathclap{s' \in \{0, 1\}}} P(1, a, s') \cdot V(s')] \label{eqn:max}
\end{align}
These are 3 equations in 3 unknowns ($V(0), V(1), m_{s_0}$). The only hiccup here is that Equation \ref{eqn:max} has a $\max$ term and so this set of equations can not be solved as normal linear equations would be. However, we can `unroll' Equation \ref{eqn:max} into 2 different equations:
\begin{align}
    V(1) &= R(1) + m_{s_0} + \gamma \sum_{\mathclap{s' \in \{0, 1\}}} P(1, 0, s') \cdot V(s'), \quad \quad \text{or} \\
    V(1) &= R(1) + \gamma \sum_{\mathclap{s' \in \{0, 1\}}} P(1, 1, s') \cdot V(s') \label{eqn:argmax-selection}
\end{align}
Each of these corresponds to evaluating the value function associated with the partial policies $s = 1 \rightarrow a = 0$ and $s = 1 \rightarrow a = 1$. Then to get the optimal policy, we can just evaluate both of the policies and choose the better of the two policies, i.e., the policy with the higher expected value $V(1)$.
In practice, we pre-compute the Whittle index and value function using the binary search and value iteration approach studied by ~\citet{qian2016restless}.
Therefore, to determine which equation is satisfied, we just use the pre-computed value functions to evaluate the expected future return of different actions, and use the one with higher value to form a set of linear equations.

This gives us a set of linear equations where Whittle index is a solution. We can therefore derive a closed-form expression of the Whittle index as a function of the transition probabilities, which is differentiable. This completes the differentiability of Whittle index. This technique is equivalent to saying that the policy does not change if we infinitesimally change the input probabilities.


\subsection{Worked Example}

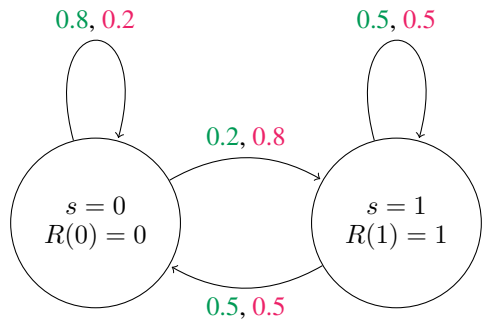
\begin{figure}[ht]
    \centering
    \begin{tikzpicture}[shorten >=1pt,node distance=4cm,on grid,auto]
      \node[state] (s_0) {\begin{tabular}{c} $s = 0$ \\ $R(0) = 0$ \end{tabular}};
      \node[state] (s_1) [right of=s_0]  {\begin{tabular}{c} $s = 1$ \\ $R(1) = 1$ \end{tabular}};
    
      \path[->]
      (s_0) edge [bend left] node {\textcolor{ForestGreen}{0.2}, \textcolor{WildStrawberry}{0.8}} (s_1)
            edge [loop above]  node {\textcolor{ForestGreen}{0.8}, \textcolor{WildStrawberry}{0.2}}  (s_0)
      (s_1) edge [bend left] node {\textcolor{ForestGreen}{0.5}, \textcolor{WildStrawberry}{0.5}} (s_0)
            edge [loop above]  node {\textcolor{ForestGreen}{0.5}, \textcolor{WildStrawberry}{0.5}}  (s_1);
    \end{tikzpicture}
    \caption{An MDP with the probabilities associated with the \textcolor{ForestGreen}{passive action $a = 0$ in red} and \textcolor{WildStrawberry}{active action $a = 1$ in green}.}
    \label{fig:example}
\end{figure}

\noindent Let us consider the concrete example in Figure \ref{fig:example} with $\gamma = 0.5$. To calculate the whittle index for state $s = 0$, we have to solve the following set of linear equations:
\begin{align*}
    V(0) &= 0 + m_{s_0} + 0.5 \cdot [0.8 V(0) + 0.2 V(1)] &  V(0) &= 0 + m_{s_0} + 0.5 \cdot [0.8 V(0) + 0.2 V(1)] \\
    V(0) &= 0 + 0.5 \cdot [0.2 V(0) + 0.8 V(1)] & V(0) &= 0 + 0.5 \cdot [0.2 V(0) + 0.8 V(1)]\\
    \textcolor{ForestGreen}{V(1)} &\textcolor{ForestGreen}{= 1 + m_{s_0} + 0.5 \cdot [0.5 V(0) + 0.5 V(1)]} & \textcolor{WildStrawberry}{V(1) } &\textcolor{WildStrawberry}{= 1 + 0.5 \cdot [0.5 V(0) + 0.5 V(1)]} \\ \\
    \textcolor{ForestGreen}{V(0)} & \textcolor{ForestGreen}{\approx 0.65, V(1) \approx 1.45, m_{s_{\mathrlap{0}}} \approx 0.25} & \textcolor{WildStrawberry}{V(0)} &\textcolor{WildStrawberry}{\approx 0.52, V(1) \approx 1.18, m_{s_{\mathrlap{0}}} \approx 0.20}
\end{align*}
Here the left set of equations corresponds to taking action $a = 0$ in state $s = 1$ and the right corresponds to taking the action $a = 1$. 
As we can see in the above calculation, given subsidy $m_{s_0}$, it is better to choose the passive action (a=0) on the left to obtain a higher expected future value $V(1)$. On the other hand, this can also be verified by precomputing the Whittle index and the value function.
Therefore, we know that the passive action in Equation~\ref{eqn:argmax-selection} leads to a higher value, where the equality holds.
Thus we can express the Whittle index as a solution to the following set of linear equations:
\begin{align*}
    V(0) &= R(0) + m_{s_0} + \gamma \sum_{\mathclap{s' \in \{0, 1\}}} P(0, 0, s') \cdot V(s') \\
    V(0) &= R(0) + \gamma \sum_{\mathclap{s' \in \{0, 1\}}} P(0, 1, s') \cdot V(s') \\
    V(1) &= R(1) + m_{s_0} + \gamma \sum_{\mathclap{s' \in \{0, 1\}}} P(1, 0, s') \cdot V(s')
\end{align*}

By solving this set of linear equation, we can express the Whittle index $m_{s_0}$ as a function of the transition probabilities. Therefore, we can apply auto-differentiation to compute the derivative $\frac{d m_{s_0}}{d P}$.

\end{document}